\documentclass[journal]{IEEEtran}
\ifCLASSINFOpdf
\else
\fi
\usepackage{times}
\usepackage{soul}
\usepackage{url}
\usepackage[hidelinks]{hyperref}
\usepackage[utf8]{inputenc}
\usepackage[small]{caption}
\usepackage{graphicx}
\usepackage{amsmath}
\usepackage{amsthm}
\usepackage{booktabs}
\usepackage{algorithm}
\usepackage{algorithmic}
\usepackage{subfigure} %需要使用的宏包
\usepackage{multirow}
\usepackage{color}

\begin{document}
\title{SpatioTemporal Focus for Skeleton-based Action Recognition}

\author{Liyu~Wu, Can~Zhang,~\IEEEmembership{Member,~IEEE,}
        and~Yuexian~Zou*,~\IEEEmembership{Senior Member,~IEEE}
        % <-this % stops a space
% \thanks{M. Shell was with the Department
% of Electrical and Computer Engineering, Georgia Institute of Technology, Atlanta,
% GA, 30332 USA e-mail: (see http://www.michaelshell.org/contact.html).}% <-this % stops a space
\thanks{L. Wu, C. Zhang and Y. Zou are with the School of Electrical and Computer Engineering, Peking
University, China (E-mail: wuliyu@pku.edu.cn; zhangcan@pku.edu.cn; zouyx@pku.edu.cn).
}% <-this % stops a space
\thanks{* \textit{Corresponding author:} Yuexian Zou.}}


\markboth{Journal of \LaTeX\ Class Files,~Vol.~14, No.~8, August~2015}%
{Shell \MakeLowercase{\textit{et al.}}: Bare Demo of IEEEtran.cls for IEEE Journals}



\maketitle

% title selection
% STF-Net: spatiotemporal focus network for skeleton-based Action Recognition
% SpatioTemporal focusing Network for skeleton-based Action Recognition
%SpatioTemporal focus for skeleton-based Action Recognition
\begin{abstract}
Graph convolutional networks (GCNs) are widely adopted in skeleton-based action recognition due to their powerful ability to model data topology. We argue that the performance of recent proposed skeleton-based action recognition methods is limited by the following factors. First, the predefined graph structures are shared throughout the network, lacking the flexibility and capacity to model the multi-grain semantic information. Second, the relations among the global joints are not fully exploited by the graph local convolution, which may lose the implicit joint relevance. For instance, actions such as running and waving are performed by the co-movement of body parts and joints, \textit{e.g.}, legs and arms, however, they are located far away in physical connection. Inspired by the recent attention mechanism, we propose a multi-grain contextual focus module, termed MCF, to capture the action associated relation information from the body joints and parts. As a result, more explainable representations for different skeleton action sequences can be obtained by MCF. In this study, we follow the common practice that the dense sample strategy of the input skeleton sequences is adopted and this brings much redundancy since number of instances has nothing to do with actions. To reduce the redundancy, a temporal discrimination focus module, termed TDF, is developed to capture the local sensitive points of the temporal dynamics. MCF and TDF are integrated into the standard GCN network to form a unified architecture, named STF-Net. It is noted that STF-Net provides the capability to capture robust movement patterns from these skeleton topology structures, based on multi-grain context aggregation and temporal dependency. Extensive experimental results show that our STF-Net significantly achieves state-of-the-art results on three challenging benchmarks NTU RGB+D 60, NTU RGB+D 120, and Kinetics-skeleton.
\end{abstract}


% \fi
\IEEEpeerreviewmaketitle

\section{Introduction}
\label{sec:intro}
% for IEEE journal papers produced under \LaTeX\ using
% IEEEtran.cls version 1.8b and later.
% % You must have at least 2 lines in the paragraph with the drop letter
% % (should never be an issue)
% I wish you the best of success.

 

\IEEEPARstart{A}{ction} recognition has become one of the research hotspots of the computer vision community due to its potential applications in many real-world scenarios, such as video surveillance, video retrieval, human-object interaction, \textit{etc}~\cite{in_GraphNet_ijcai2020,yang2021rr_net}. Currently, the skeleton-based action recognition task has attracted much attention due to the development of depth sensors~\cite{Zhu_Lan_Xing_Zeng_Li_Shen_Xie_2016_depth_sensor} and pose estimation algorithms~\cite{Cao_2017_CVPR_pose_estimation,Sun_2019_CVPR_pose_estimation,openpose}. The human skeleton is usually represented by either locating 2D or 3D spatial coordinates of joints and  has several characteristics. First, the skeleton is the abstract representation of human posture and behavior. Intuitively, we can recognize human action by observing only the motion of joints even without appearance information. Second, compared to the raw RGB modality, skeleton data is considered as a more robust representation (e.g. background clutter, lighting conditions, clothing) for human action dynamics. Third, it is also computationally efficient because of the low dimensional representation and compact data presentation, making it possible to design more lightweight and powerful models. Meanwhile, skeleton modality is also complementary to the RGB-based action recognition \cite{skeleton_modality}. Due to the mentioned advantages of the skeleton modality, we focus on skeleton-based action recognition in this work.

\begin{figure}[t]
   \centering
   \includegraphics[width=8.5cm]{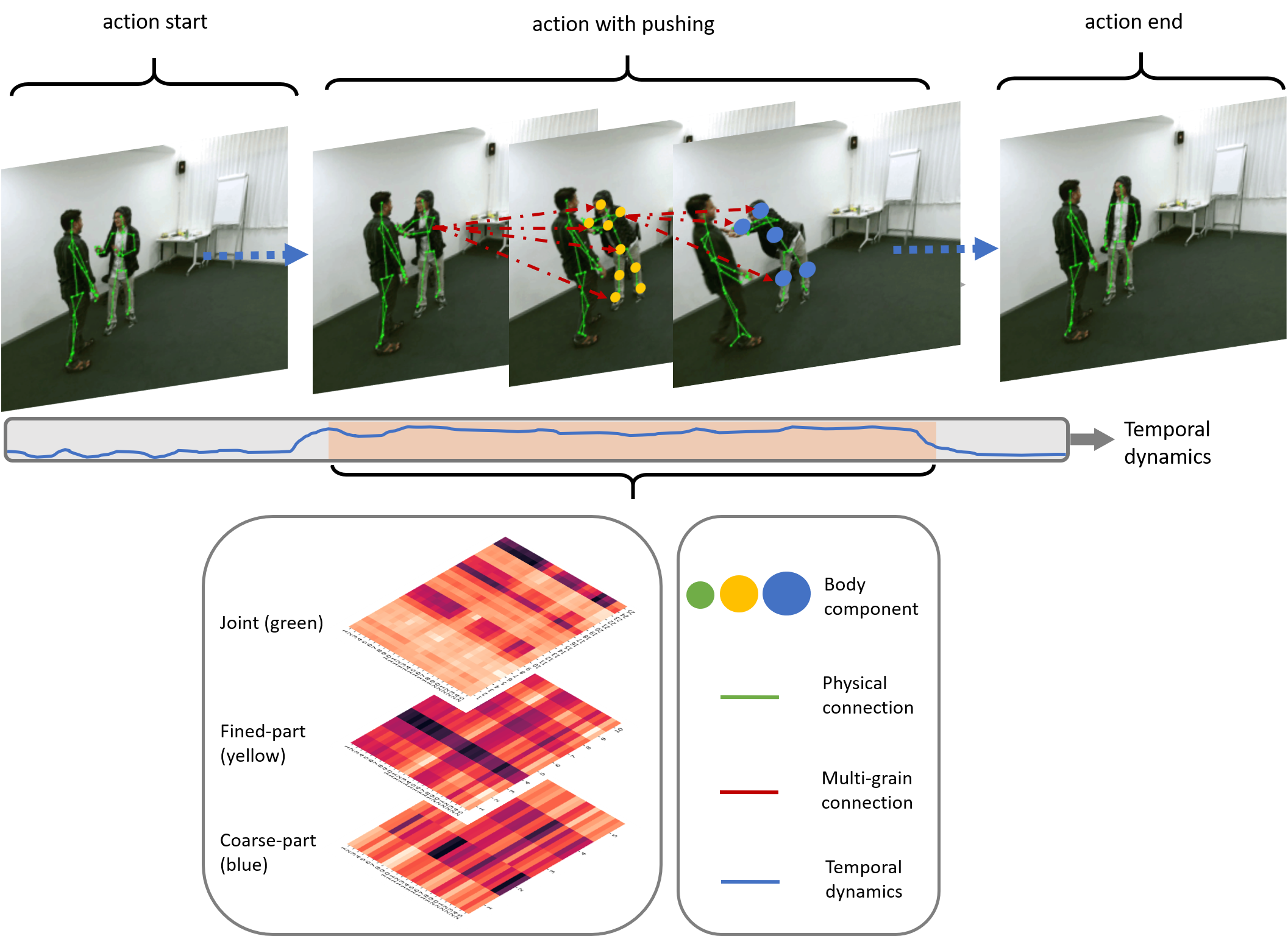}
   \vspace{1em}
%   Red arrows denote the integration of contextual information from different grains in the spatial dimension. green lines denote the physical connection inherent in the raw skeleton.
   \caption{The illustration of contextual information in skeleton sequence. The heatmaps show the correlation among different scales (joint, fined-part, and coarse-part) of body components. The green line denotes the physical connection of the skeleton and the red line denotes integration of context information along spatial dimension. The blue curve denotes the discrimination of the temporal dynamics. The temporal dynamics marked in orange color represent the main part of the action sequence. We argue that finding the discriminative spatiotemporal information is critical for learning reliable topology.}
\label{fig:multigrain}
\end{figure}
% \begin{figure}[t]
% \centering
% \subfigure[]{\includegraphics[width=8cm]{IEEEtran/figures/pic1_1.png}}
% \\ %换行
% \centering
% \subfigure[]{\includegraphics[width=8cm]{IEEEtran/figures/pic1_2.png}}
% \\
% \caption{The contextual information of skeleton sequence. Red arrows denote the integration of contextual information from different grains in the spatial dimension. Gray lines denote the physical connection inherent in the raw skeleton. The green curve denotes the discrimination of the temporal dynamic.} %图片标题
% \label{fig:multigrain}
% \end{figure}

Skeleton-based action recognition aims to recognize the actions in skeleton sequences that contains a time series of human joint coordinates. Recently, deep learning is widely explored to model the spatiotemporal skeleton sequence. Previous works adopt Recurrent Neural Networks (RNNs)~\cite{Ke_2017_CVPR_RNN,Kim_2017_CVPR_Workshops_RNN,Du_2015_CVPR_RNN,ST_LSTM_RNN}, Convolutional Neural Networks (CNNs)~\cite{ICMEW_2017_CNN,IJCAI_2018_CNN,Goated_CNN}, where the coordinates of joints are represented as vector sequences and resized to pseudo-images, respectively. Nevertheless, none of the mentioned methods explicitly explore the topology information of human skeleton data. Graph Convolutional Network (GCN) methods have gained impressive successes of the skeleton-based action recognition due to the capability of exploiting the topological structure of the skeleton. For example, Spatial-Temporal Graph Convolutional Networks (ST-GCN)~\cite{stgcn2018aaai} first introduces the GCN and uses the adjacent matrix to aggregate the physical structure of the skeleton data. The adjacent matrix represents the physical connection of the human body and aims to propagate the information between subgraphs (the matrix is usually decoupled into several sub adjacent matrixes) via the shared nodes. After that, increasing studies are reported based on GCN models\cite{2sagcn2019cvpr,Li_2019_CVPR_ASGCN,shi2019skeleton,Liu_2020_CVPR_G3D,zhang2020context,song2020richly,yang2018action,wang2021multi,miao2021central,wu2021graph2net}. 

Although GCN-based methods have achieved impressive performance, there are still some issues to be addressed in modeling skeleton sequences. (1) In most of the works, the adjacent matrix is usually heuristically predefined and represents only the physical structure of the human body. However, such heuristical architecture may be sub-optimal in action recognition due to the lack of flexibility~\cite{2sagcn2019cvpr}. (2) The graph convolution only works in a local manner. Generally, the receptive field can be increased by stacking multiple convolutional layers, but this may dilute the importance of the target joints. (3) Moreover, accurate action recognition requires selectively focusing on some relevant body parts while ignoring others (\textit{e.g.}, hands for clapping and arms for waving). Intuitively, body joints and body parts contain individual semantic information. As shown in Fig.~\ref{fig:multigrain}, the human arms are crucial in recognizing the action with pushing while they are less critical during the action start and end procedure. Besides, actions such as running and waving are performed by the co-movement of body parts, \textit{e.g.}, legs and arms. We argue that multi-grain contextual information is critical for learning reliable topology. The interpretability issues of conventional models still lack considerations in current studies. (4) A intuitive idea is to focus on the temporal dimension of the sequences. However, not all video frames are well-related to the recognition target, some may harm the dynamics of the skeleton sequence. As shown in Fig.~\ref{fig:multigrain}, a complete sequence includes the duration of starting, actionness, ending~\cite{DBLP:conf/eccv/LinZSWY18} where actionness may also contain different sub-actions. On the other hand, the widely-used dense sampling strategy may bring heavy redundancy, since adjacent video frames often have high similarity in the skeleton sequence. 
% duplicate
% The dense sampling strategy causes inter-frame redundancy, and adjacent video frames tend to have great similarity in skeleton sequences.

To tackle these problems, we propose a novel skeleton-based GCN framework, termed spatiotemporal focus network (STF-Net). STF-Net consists of two modules, multi-grain contextual focus module (MCF) and temporal discrimination focus module (TDF), which can be instantiated with existing GCN at a small extra computational cost. Specifically, MCF is proposed to capture the contextual information among body joints and body parts via asymmetric non-local operation \cite{zhu2019asymmetric} in the spatial domain, where we split the human body into different components heuristically. For temporal modeling, TDF is designed to enhance discriminative temporal information from a local view by extending the method of squeeze-and-excitation~\cite{hu2018squeeze}. In this way, it is able to capture the distinctive dynamics properties of an action instance.

Following the previous work \cite{2sagcn2019cvpr}, in addition to the graph convolution operation, STF-Net also adopt the trainable adjacent matrix, graph mask. Under the data-driven manner, the trainable masked matrix will be more flexible in learning different actions and more individualized for the information contained in different layers.

In a nutshell, our main contributions can be summarized as follows: 
\begin{itemize}
\item We propose a novel GCN network, termed STF-Net, to explicitly model the multi-grain contextual information among body joints and body parts for efficient skeleton-based action recognition.
\item We propose two key modules, multi-grain contextual focus (MCF) and temporal discrimination focus (TDF) for STF-Net. MCF that captures multi-grain relations among joints and parts in the spatial domain based on asymmetric non-local, and TDF that perceives the distinctive local dynamics properties of an action instance by extending the method of squeeze-and-excitation.
\item On three large-scale benchmarks for skeleton-based action recognition, including NTU-RGBD-60, NTU-RGBD-120 and Kinetics-skeleton, our proposed STF-Net achieves state-of-the-art performance, demonstrating the effectiveness of our method. 
\end{itemize}

\section{Related Work}

In skeleton-based action recognition, deep learning methods are proved to be more effective than conventional hand-crafted methods. We will briefly introduce the related work in this section, including skeleton-based action recognition, graph convolutional networks, and multi-scale methods.

\subsection{Video Representation Learning}

Video representation learning is a core topic in computer vision. In the early stage, lots of traditional methods design hand-crafted features~\cite{le2011learning,sadanand2012action} to represent video data. Recently, the development of video understanding has benefited greatly from deep learning methods.  A series of CNNs-based methods were proposed to learn spatiotemporal representations in video recognition. Specifically, TSN~\cite{wang2018temporal} used the sparsely sampled frames from the whole video to learn the long-range information by aggregating scores after the last fully-connected layer. TSM~\cite{lin2019tsm} shifted the channels along the temporal dimension in an efficient way, which yields a good performance with 2D CNNs. By a simple extension from the spatial domain to the spatiotemporal domain, 3D convolution was proposed to capture the appearance and motion information simultaneously in an unified network, such as C3D~\cite{tran2015learning} and I3D~\cite{carreira2017quo}. Attention mechanism is also widely used in action recognition, \textit{e.g.,} method like~\cite{Wang_2018_CVPR_nonlocal} utilizes the self-attention mechanism to capture the global spatiotemporal features in videos. Besides, combining motion representations with RGB ones will generally boosts the video recognition performance. Effective motion representations including: optical flow~\cite{wang2018temporal,carreira2017quo}, optical flow guided feature (OFF)~\cite{sun2018optical}, PA~\cite{zhang2019pan} and RGB difference~\cite{wang2018temporal}, \textit{etc}.

Compared to the above-mentioned video representations , the human skeleton is a more robust representation since it eliminates the influences of background clutter, lighting interference, clothing variance in real-world videos.

\subsection{Skeleton-Based Action Recognition}
The human skeleton is easy to obtain through  highly accurate depth sensors~\cite{Zhu_Lan_Xing_Zeng_Li_Shen_Xie_2016_depth_sensor} or pose estimation algorithms~\cite{Sun_2019_CVPR_pose_estimation}. In addition, compared with the RGB modality, human skeleton has strong robustness to the illumination and scene variation. Early approaches~\cite{DBLP:conf/cvpr/DuWW15,shahroudy2016ntu} focus on hand-crafted features, which ignore the semantic relationship among human joints. However, deep learning methods have been prosperous in the skeleton-based field, indicating the importance of the semantic human skeleton for action predictions. RNN is a straightforward way to model the skeleton data as a sequence of the coordinate vectors, where each vector represents a human body joint. RNNs construct temporal information sequentially, while CNNs  encode the spatiotemporal information jointly. A common practice of CNN-based methods is to construct the skeleton as a pseudo-image based on the predefined transformation regulations. For instance~\cite{DBLP:conf/avss/CaetanoSBSS19} directly transforms the coordinates to a pseudo image, typically generating a shape of $K \times T$ 2D input, where $K$ is the number of joints and $T$ is the temporal length. However, these methods either ignore the information of skeleton topology among joints or suffer from complicated design processes, which make these methods not as competitive as graph-based methods on the popular benchmarks.

In contrast, inspired by the physical topology of the human body, graph-based methods have recently achieved impressive performance. ST-GCN~\cite{stgcn2018aaai} firstly introduces graph convolution operation and temporal convolution to model the spatiotemporal dimensions of the skeleton simultaneously. To make the graph topology more flexible, AGCN~\cite{2sagcn2019cvpr} proposes the adaptive adjacent matrix to adjust better graph for the skeleton. DGNN~\cite{shi2019skeleton} represents the skeleton as a directed acyclic graph, which constructs the joint and edge information  as well as the bone features. It updates the joint and bone features through an alternating spatial aggregation scheme simultaneously. AS-GCN~\cite{Li_2019_CVPR_ASGCN} infers A-links from input data to capture actional dependencies and uses multi-scale adjacent matrix for presenting structural joints. With the popularity of Transformer recently, ST-TR~\cite{DBLP:conf/icpr/PlizzariCM20} proposes a novel network which models dependencies between joints using the Transformer self-attention operator.

\subsection{Graph Convolutional Network}

GCN has attracted more attention due to its ability of modeling non-Euclidean structure data, such as graph structure. Generally, there are two mainstreams identifying the principle of constructing GCNs. 1) Spectral perspective is a way that utilizes the eigenvalues and eigenvectors of the graph Laplace matrices \cite{bruna2014spectral,defferrard2017convolutional_spectral,henaff2015deep_spectral}. Graph convolution is performed in the frequency domain by using graph Fourier transform\cite{Shuman_2013_spectral}. 2) Spatial perspective is a straightforward way that directly designs convolution by weighted summation of vertices on spatial domain\cite{niepert2016learning_spatail}. Following this stream, many works have applied spatial GCNs to the computer vision task. Our work constructs the CNN filters on the graph and follows the principle of the spatial perspective.

\subsection{Multi-Scale Methods}

Multi-scale feature representation has been widely used in computer vision and is proved effective. One of the famous applications is the feature pyramid network \cite{lin2017feature_pyramid}. FPN combines shallow features and deep features (deep high-level semantics, low resolution; latent low-level semantics, high resolution) to deal with the shortcoming of object detection with multi-scale  characteristics of object detection. Other works relate to the multi-scale features like \cite{cai2016unified_fast_detection, chen2018biglittle, newell2016stacked_hourglass} do contribute in detection or recognition fields. Recently, Cross-ViT \cite{chen2021crossvit} proposes a dual-branch (one large patch and another small patch) transformer to combine the patches of different sizes to identify multi-scale objects. The two branches are merged through multiple attentions. bLVNet-TAM \cite{fan2019blvnet} follows~\cite{chen2018biglittle} the two-branch architecture which input more frames under different resolutions. SlowFast \cite{feichtenhofer2019slowfast} processes a video at both slow and fast sample rates. A slow pathway running at a low frame rate captures spatial semantic information. While a fast pathway runs at a high frame rate to capture motion information with fine time resolution. Back to the skeleton-based action recognition,  GNN methods in multi-scale spatial dimension have also been proposed to capture semantics from non-local neighbors. \cite{stgcn2018aaai,DBLP:conf/iclr/LiaoZUZ19} use higher order polynomials of the graph adjacency matrix to aggregate features from long-range neighbor nodes. G3D \cite{Liu_2020_CVPR_G3D} proposed multi-scale aggregation scheme that disentangles the importance of nodes in different neighborhoods for effective long-range modeling and address the biased weighting problem.

\section{Method}

In this section, first we review the background of GCN in skeleton-based action recognition and how to construct the skeleton graph. We illustrate the details of the designed multi-grain contextual focus module (MCF) and the temporal discrimination focus module (TDF). we integrate all the components and propose spatiotemporal focus network (STF-Net) for skeleton-based action recognition.

\begin{figure*}[t]
   \centering
   \includegraphics[width=18cm]{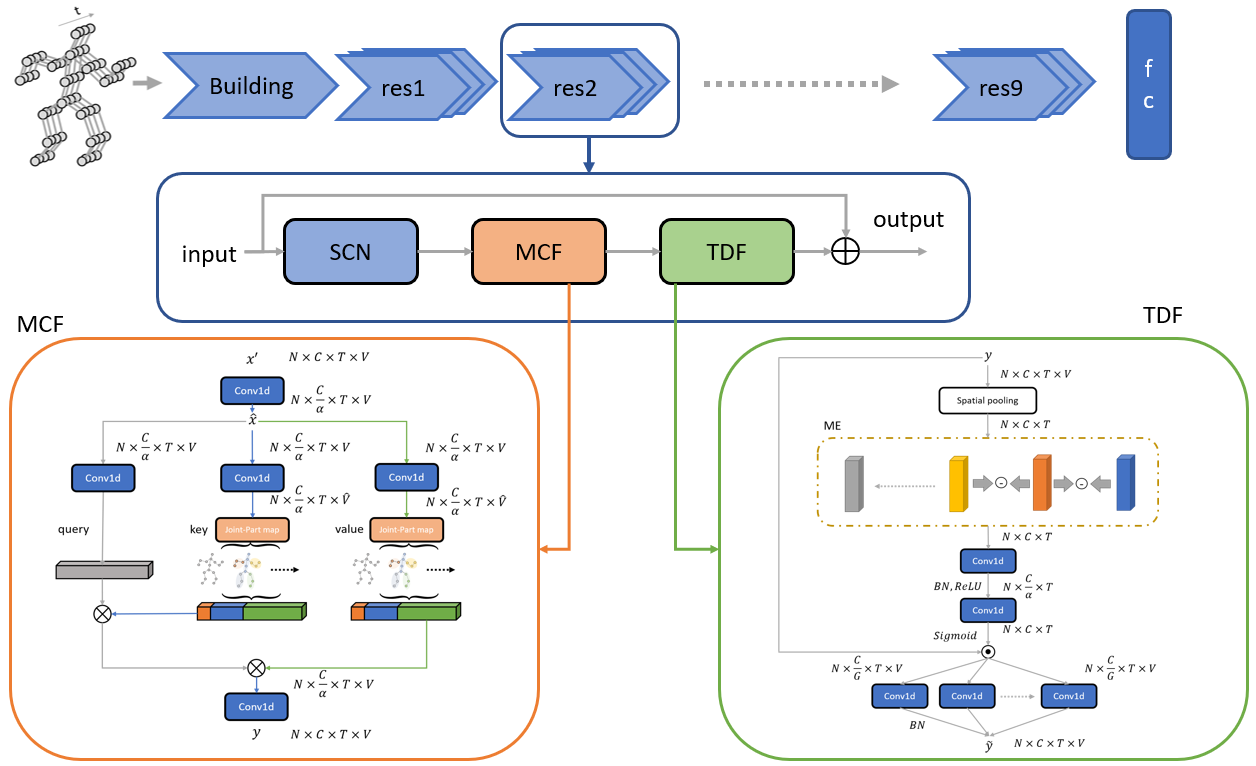}
   \vspace{1em}
   \caption{Architecture of our proposed spatiotemporal focus network (STF-Net). STF-Net contains nine res-layer following building block, each res-layer is simply formed by three parts, which are SCN and two proposed modules, MCF and TDF. MCF aims to capture the multi-grain spatial features and TDF trys to percieve the local sensitive temporal dynamics. Finally, the network followed by a fully connecting layer with Softmax.}
\label{fig:mcnet}
\end{figure*}

% \section{Graph construction}
\subsection{A Formulation of Spatial Graph Convolution}
Following ST-GCN \cite{stgcn2018aaai}, a human skeleton graph is given as $G(V,E)$ , where $V=\{v_1, v_2, ..., v_V\}$ is the set of $V$ body joints and $E=\{e_1, e_2, ..., e_M\}$ is a set of $M$ bones or edges. Let $A \in \{0, 1\}^{V\times{V}}$ be an adjacent matrix of the skeleton graph, where $A_{i,j} = 1$ denotes the i-th joint and j-th joint are connected, otherwise $A_{i,j} = 0$. In practice, we set $\hat{A} = A + I$, where $I$ is the identity matrix. The adjacent matrix $\hat{A}$ fully represents the graph structure. In addition, we denote $D \in \mathbf{R}^{V\times{V}}$ as the diagonal degree matrix of $\hat{A}$, where $D_{i,i}= \sum_{i}^{V} \hat{A}_{i,i}$. Suppose we have the input skeleton sequence $x \in \mathbf{R}^{N\times{C}\times{T}\times{V}}$ ($N$ batchsize, $C$ channel, $T$ frame, and $V$ joints), the standard graph convolution is typically described in the following formulation:
\begin{equation}
\label{eq.1}
    x' = \sigma(\sum_{i}^{K} G_{i} x W_i), x' \in R^{N\times{C}'\times{T}\times{V}}
\end{equation}
where$\quad G_{i} =  D_{i}^{-\frac{1}{2}} \hat{A}_{i} D_{i}^{\frac{1}{2}}$. $G_i$ describes the normalization of each adjacent matrix $\hat{A}_{i}$, and $\sigma$ is an activation function. $K$ denotes the number of spatial configurations which can be seen as the number of different adjacent sub-matrices with index $i$. Each adjacent sub-matrix $\hat{A}_i$ aims to capture the different location information of the skeleton graph. In detail, we part three types of adjacent sub-matrices, including (1) the root joint itself, (2) the centripetal group, which are closer to the body center than current root, and (3) the centrifugal group. $W_i$ denotes the trainable weights for each partition group.

However, there are some discussions with the standard graph convolution operation. First, The adjacent matrix $A$ is predefined and represents the physical structure of the skeleton, which may have less flexibility. Second, it is reasonable that we will need to learn the strength of the connection between each joint pair according to the different action instances. 

Basing on \cite{2sagcn2019cvpr}, we reformulate Eq.\ref{eq.2} as:
\begin{equation}
\label{eq.2}
\begin{aligned}
    x' = \sigma(\sum_{i}^{K} (G_i + G_{a}) xW_i), x' \in R^{N\times{C}'\times{T}\times{V}}
\end{aligned}
\end{equation}
In the formulation of Eq.\ref{eq.2}, $G$ is as same as in the Eq.\ref{eq.1}, while $G_{a}$ is a learnable masked matrix. $G_{a}$ is a kind of attention mechanisms, that represents how strong the existing connections are locally but also learns the extra connections between the joints in the global term, which is more flexible to the fixed structure of $\hat{A}$. In our experiments, $G_{a}$ is initialized with $G_i$. The graph convolution method that we use to construct the physical structure of the skeleton is called spatial convolution (SCN) in our proposed network, as shown in Fig.~\ref{fig:mcnet}.

% \section{Method}

\subsection{Multi-grain Contextual Focus module (MCF)}
The multi-grain spatial contextual module aims to enrich the traditional local graph convolution so that the centered vertex will focus on its surrounding neighbors and be aware of distant vertices. As mentioned above, body components in different granularities may contain their own individual semantic information, therefore, besides considering the contextual information among each joint, we also explore the diffusion across different granularity.

Inspired by the operation of self-attention and non-local which are potent to capture the long-range dependencies and could be used to determine whether there is a connection between two body components and how strong the connection is. The similarity of two body components could be measured by the normalized embedded Gaussian function as:

\begin{equation}
\label{eq:3}
    f(v_i, v_j) = \frac{e^{\theta(v_i)^T\phi(v_j)}}{\sum_{j=1}^{V} e^{\theta(v_i)^T\phi(v_j)}} ,
\end{equation}
where $V$ is the number of body components or joints. In order to measure the similarity across different granularity, we design a new module called MCF, as shown in Fig.~\ref{fig:mcnet}. Given a input feature  obtained from SCN, we first pass the first $x'$ obtained from SCN through the  $1\times{1}$ convolution to reduce the channel dimension $\hat{x} \in  R^{N\times{\frac{C}{\alpha}}\times{T}\times{V}}$  for efficient concern, where N, C, T, and V denote the dimension of the batchsize, channel, frame, and joint respectively.Three $1\times{1}$ convolutions transform $\hat{x}$ into linear embedding as:

\begin{equation}
\begin{aligned}
\gamma(x)& = W_{\gamma}\hat{x}, \gamma(x) \in R^{N\times{\frac{C}{\alpha}}\times{T}\times{V}} \\
\phi(x)& = W_{\phi}\hat{x}, \phi(x) \in R^{N\times{\frac{C}{\alpha}}\times{T}\times{V}} \\ 
\theta(x)& = W_{\theta}\hat{x}, \theta(x) \in R^{N\times{\frac{C}{\alpha}}\times{T}\times{V}} \\
\end{aligned}
\end{equation}
where $\gamma$, $\phi$, and $\theta$ represent $Q$, $K$, and $V$, which are formulated in the self-attention form~\cite{vaswani2017attention}. To enable information diffusion across multi granularity, we first map $K$ and $V$ ,the joint-level feature, into part-level feature, for instances $\gamma(x)' \in R^{N\times{\frac{C}{\alpha}\times{T}\times{\hat{V}}}} $, where $\hat{V}$ is defined under heuristic joint-part mapping $p_i$, where $i$ is the index of the mapping function. So we have $p_i(\phi(x)) \in R^{N\times{\frac{C}{\alpha}}\times{T}\times{\hat{V}_i}}$ and $p_i(\theta(x)) \in R^{N\times{\frac{C}{\alpha}}\times{T}\times{\hat{V}_i}}$
After data permutation and reshape operation, a straightforward pipeline for measuring similarity across granularity is given as:
\begin{equation}
\begin{aligned}
R^{N\times{V\times{\frac{C}{\alpha}}T}}\times{R^{N\times{\frac{C}{\alpha}}T\times{\hat{V}}}} &\xrightarrow{} R^{N\times{V}\times{\hat{V}}}\times{R^{N\times{\hat{V}}\times{\frac{C}{\alpha}T}}} \\
&\xrightarrow{} R^{N\times{\frac{C}{\alpha}}\times{T}\times{V}}
\label{eq:5}
\end{aligned}
\end{equation}
corresponding as:
\begin{equation}
\gamma(x)\times{p_{i}(\phi(x))} \xrightarrow{} Softmax(M) \times{{p_{i}(\theta(x))}} \xrightarrow{} \tilde{x_{i}}
\label{eq:6}
\end{equation}
where M is the similarity matrix generated by $Q$ multiplying $K$. This is straightforward to see that in Eq.~\ref{eq:5} and Eq.~\ref{eq:6}, each joint could be able to integrate with a different scale of parts.

\noindent\textbf{Weighted sum fusion.} After obtaining all the contextual information $\tilde{x_{i}}$, it is straightforward to adjust the importance among each contextual information since each action requires different granularity of body parts. In the experiments, we find out weight sum is the best choice to obtain higher accuracy.
\begin{equation}
y' = \sum_{i\in P} \omega_{i}\tilde{x_{i}},
\end{equation}
where we let the network learn the parameter $w_i$ automatically during the training process. Finally, we use $1\times{1}$ convolution to recover the channel dimension of $y'$ to $y \in R^{N\times{C}'\times{T}\times{V}}$.

We insert our MCF into the res-layer4 to res-layer9. The reason is that the receptive field at the bottom is small, which limits the ability to learn graph topology from different samples. The information in the top layer is more semantic and variable, and the graphic topology is expected to be more characteristic.

In this way, the spatial contextual information is fully considered through MCF. However, compared to images that only contain spatial information, videos naturally carry out rich temporal information, they can also contain redundant temporal information between neighboring frames and result in less dynamic. Therefore we propose TDF to capture the sensitive temporal dynamics.

\subsection{Temporal Discrimination Focus module (TDF)}
As we discussed in Sec.1, skeleton sequence typically exhibits the complex temporal dynamics and redundancy caused by different action instances. To tackle this problem, TDF tries to capture the discriminative dynamic for each action instance. The detailed architecture of TDF is shown in Fig.~\ref{fig:mcnet}.

Since Our TDF only focus on temporal modeling instead of the spatial pattern, after the input skeleton sequence $y\in R^{N\times{C}\times{T}\times{V}}$ outputs from MCF we first adopt a global spatial pooling to squeeze the feature map as follows:

\begin{equation}
    \hat{y}_{n, c, t} = F_v(y_{n, c, t, v}) = \frac{1}{V}\sum_{v=1}^{V}y_{n, c, t, v},
\end{equation}
where $n$, $c$, $t$, and $v$ represent the index of the batchsize, channel, temporal (frames), and spatial (joints) dimensions. $F_v$ is the global average pooling function, that aggregates the spatial information. Before we start to excite the temporal information, we first give a try on exploiting the motion saliency between adjacent frames by computing the difference between these two frames, ME in TDF as shown in \ref{fig:mcnet}.

1D temporal convolutions are applied on $\hat{x}_{c, t}$ to capture the dependencies in temporal and channel dimensions. The temporal excitation process can be written as:  
\begin{equation}
    \tilde{y}_{n, c, t} = F_2(\sigma(F_1(\hat{y}_{n, c, t}, W_{1})), W_2),
\end{equation}
where $F_1$ and $F_2$ represent Conv1D and $\sigma$ is activation function ReLU. To reduce the model complexity, the first Conv1D followed by BN reduces the number of channels by weight matrix $W_1\in R^{C\times{\frac{C}{\gamma}}}$, and 
$W_2\in R^{\frac{C}{\gamma}\times C}$ aims to recover the channel dimension of $\tilde{y}_{n, c, t} $ to be same as input $\hat{y}_{n, c, t} $. The difference between the temporal excitation in TDF and SENet is that SENet only learns weights for each channel of feature maps, while TDF extends temporal dimension to excited function to characterize temporal information for channel-wise features.

Most existing works perform temporal modeling using temporal convolutions with a fixed kernel size, there are also some other works such as G3D~\cite{Liu_2020_CVPR_G3D} uses multiple dilated temporal convolutions or STIGCN~\cite{huang2020spatio} uses  dedicatedly designed  kernel size to capture multis-scale information. Inspired by ResNeXt~\cite{xie2017aggregated}, the temporal Conv1D in TDF are designed that the high-dimensional convolution layers are grouped into multiple identical convolution layers (Inception is each different convolution layer), then convolution operation is carried out, and finally these convolution layers are fused. Each of group Conv1D called cardinality learning the different pattern of temporal dynamics. In the experiments, we set $\frac{C}{4}$ cardinalities, where 4 is the reduction constant.

\subsection{Spatiotemporal Focus Network (STF-Net)}

To have a head-to-head comparison with the state-of-the-art methods, we construct out spatiotemporal focus network based on the same backbone (ST-GCN~\cite{stgcn2018aaai}). As shown in Fig.~\ref{fig:mcnet}, the overall architecture of the network is the stack of these blocks. There are a total 9 blocks following a beginning block that extends the channel dimension from 3 to 64 but does not contain MCF and TDF. The numbers of the output channels for each block are 64, 64, 64, 128, 128, 128, 256, 256 and 256. Each block contains spatial convolution (SCN), multi-grain contextual focus module (MCF), and temporal discrimination focus module (TDF) sequentially.

\subsection{Multi-stream Fusion}

Following~\cite{shi2020skeleton}, the input features after various data preprocessing are divided into three classes: 1) joint positions; 2) bone features; 3) motion difference of joint and bone.

Given the original 3D coordinate $x \in R^{C\times{T}\times{V}}$, where $C$, $T$, and $V$ denote the coordinate, frame and joint, respectively. Each bone feature is represented as the direction from source joints to target joints defined by which is closer to the center of gravity, for instance, the bone feature in frame $t$ is $b_{i,j,t} = (x_{i,t} - x_{j,t}, y_{i,t} - y_{j,t}, z_{i,t} - z_{j,t})$. As for motion difference, it is calculated as the difference between the same joints or bones between two consecutive frames. For instance, given two consecutive frames $t$ and $t+1$, the motion representation can be modeled by the difference $m_{i,t,t+1} = (x_{i,t+1} - x_{i,t}, y_{i,t+1} - y_{i,t}, z_{i,t+1} - z_{i,t})$.

The four modalities (joints, bones, and their motions) are formulated by four streams separately. The predicted scores from each stream are fused using weighted summation to obtain the action scores and predict the action label.

\begin{table}[t]
\caption{Comparison of the state-of-the-art methods on the NTU RGB+D 60 dataset in terms
of the Top-1 accuracy. $\dagger$ denotes our implemented results based on their released codes on GitHub.}
\centering
\begin{tabular}{lllll}
\toprule
Model        & Conf.   & Param(M) & X-Sub & X-View \\ \midrule\midrule
TCN~\cite{DBLP:conf/cvpr/KimR17}          & CVPRW17 & -        & 74.3  & 83.7   \\
VA-fusion~\cite{DBLP:journals/pami/ZhangLXZXZ19}    & TPAMI19 & 24.60    & 89.4  & 95.0   \\ \midrule\midrule
ST-GCN~\cite{stgcn2018aaai}       & AAAI18  & 3.1$^\dagger$      & 81.5  & 88.3   \\
SR-TSL~\cite{DBLP:conf/aaai/YanXL18}       & ECCV18  & 19.07$^\dagger$    & 84.8  & 92.4   \\
GR-GCN~\cite{DBLP:conf/mm/Gao0T0G19}       & MM18    & -        & 87.5  & 94.3   \\
RA-GCN~\cite{song2020richly}       & TCSVT20 & 2.03     & 85.9  & 93.5   \\
Js CDGC~\cite{miao2021central}	 & TCSVT21 & -     & 88.0  & 94.9 \\
Graph2Net\cite{wu2021graph2net} 	& TCSVT21 & -     & 88.1  & 95.2 \\
AS-GCN~\cite{Li_2019_CVPR_ASGCN}       & CVPR19  & 6.99$^\dagger$     & 86.8  & 94.2   \\
2s AGCN~\cite{2sagcn2019cvpr}      & CVPR19  & 6.94$^\dagger$     & 88.5  & 95.1   \\
2s AGC-LSTM~\cite{DBLP:conf/cvpr/SiC0WT19}  & CVPR19  & 22.89$^\dagger$    & 89.2  & 95.0   \\
4s DGNN~\cite{shi2019skeleton}    & CVPR19    & 26.24    & 89.9  & 96.1    \\
4s MS-AAGCN~\cite{shi2020skeleton}  & TIP20   & 15.65$^\dagger$    & 90.0  & 96.2   \\
PL-GCN~\cite{DBLP:conf/aaai/HuangHOW20}       & AAAI20  & 20.7$^\dagger$     & 89.2  & 95.0   \\
2s NAS-GCN~\cite{DBLP:conf/aaai/PengHCZ20}   & AAAI20  & 13.1$^\dagger$     & 89.4  & 95.7   \\
CA-GCN~\cite{zhang2020context}       & CVPR20  & -        & 83.5  & 91.4   \\
4s shift-GCN\cite{DBLP:conf/cvpr/Cheng0HC0L20} & CVPR20  & 3.04$^\dagger$     & 90.7  & 96.5   \\
Js MS-G3D~\cite{Liu_2020_CVPR_G3D}    & CVPR20  & 3.2      & 89.4  & 95.0   \\
SGN~\cite{DBLP:conf/cvpr/0005LZXXZ20}          & CVPR20  & 1.8      & 86.6  & 93.4   \\
PA-ResGCN~\cite{DBLP:conf/mm/Song0SW20}    & MM2020  & 3.64     & 90.9  & 96.0   \\
TS-SAN~\cite{DBLP:conf/wacv/ChoM0F20}       & WACV20  & -        & 87.2  & 92.7   \\
2s ST-TR~\cite{DBLP:conf/icpr/PlizzariCM20}     & ICPR21  & -        & 90.3  & 96.3   \\ \midrule\midrule
Js STF-Net   & ours    & 1.7      & 88.8  & 95.0   \\
Bs STF-Net   & ours    & 1.7      & 89.1  & 95.1   \\
2s STF-Net   & ours    & 3.4      & 90.8  & 96.2   \\
4s STF-Net   & ours    & 6.8      & 91.1  & 96.5   \\ \bottomrule
\end{tabular}
\label{table:ntu60}
\end{table}

\begin{figure}[t]
   \centering
   \includegraphics[width=8cm]{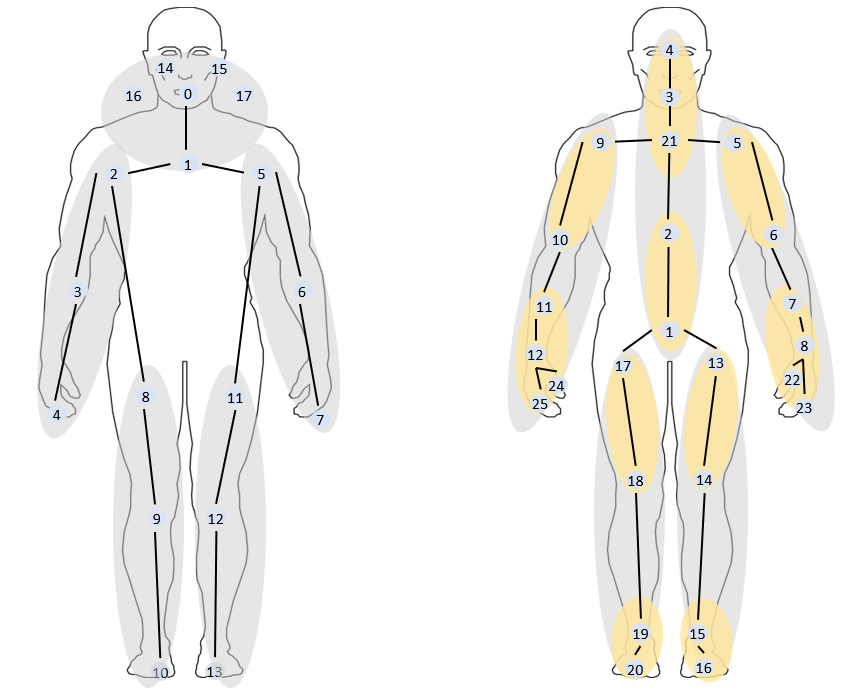}
   \vspace{1em}
   \caption{The left shows the location of the joint label of the Kinetics-Skeleton dataset and the right shows the location of the NTU-RGB+D dataset. The region of different colors represents the partition we have done in skeleton data. We split skeleton data into three different scales: joint, fined-part, and coarse-part correspond to blue node, yellow region, and gray region in NTU-RGB+D dataset. And two different scales in Kinetics-skeleton. The split skeleton data heuristically according to the body parts of humans.}
\label{fig:skeleton}
\end{figure}

\begin{figure*}[t]
   \centering
   \includegraphics[width=18cm]{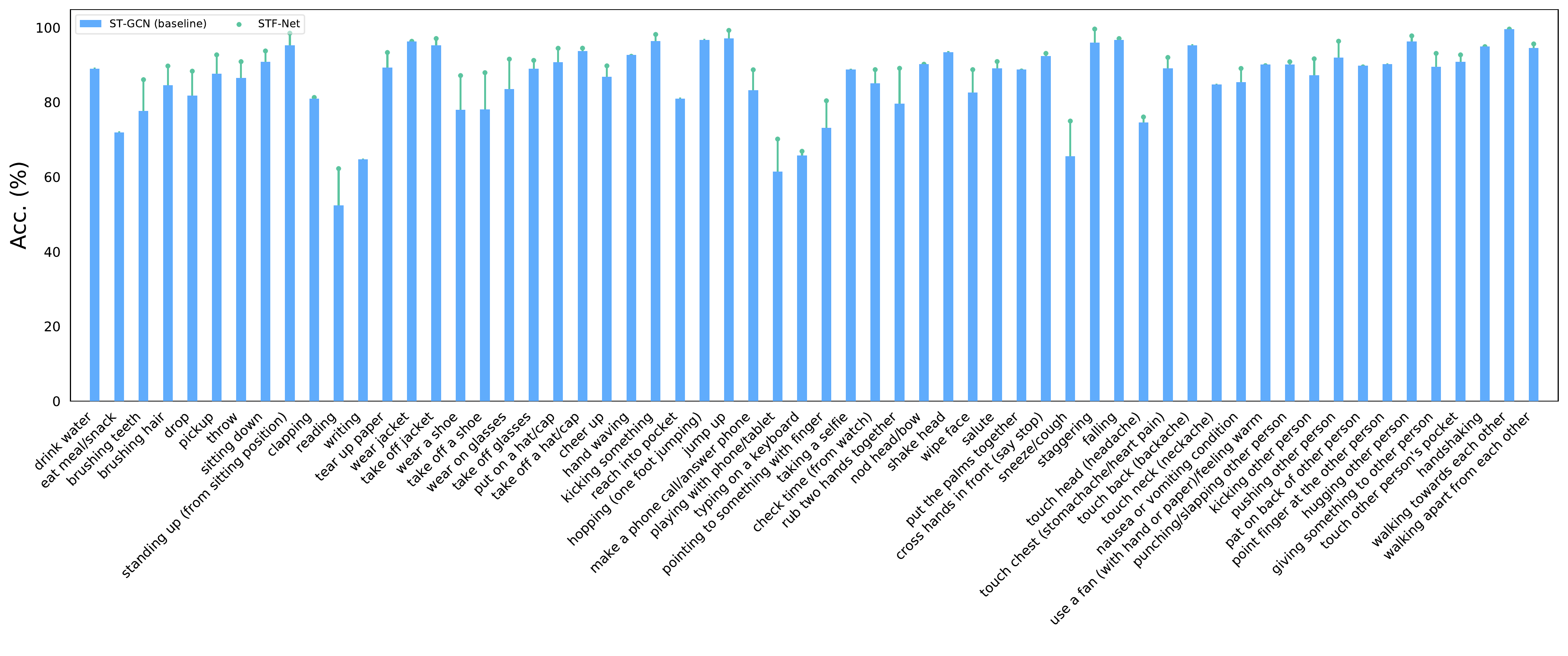}
   \caption{Classification accuracy comparison of each class on NTU-RGB+D 60 in X-sub benchmark between STF-Net and ST-GCN}
\label{fig:acc}
\end{figure*}

\section{Experimental Results}
\subsection{Datasets}
\noindent \textbf{NTU-RGB+D~\cite{shahroudy2016ntu}} is one of the most widely used datasets for human skeleton action recognition. It contains 56880 skeleton clips, with 60 action classes, including 40 types of daily actions, 9 types of health-related actions ,and 11 types of interaction actions. These clips were captured in the lab environment from three camera views. The annotations provide the 3D location (x, y, z) of each joint in the camera coordinate system. There are 25 joints per-subject. Each clip is guaranteed to contain at most 2 subjects. The authors of this dataset introduce two benchmarks: \textbf{1) Cross-subject (X-sub)} contains 40,320 clips from 20 subjects are used for training according to the ID of the subjects, and the rest of 1650 clips for evaluation. \textbf{2) Cross-view (X-view)} contains 37,920 clips captured from camera 2 and 3 are used for training and 18,960 clips from camera 1 for evaluation. According to \cite{shahroudy2016ntu}, there are 302 bad cases that should be ignored.

\noindent \textbf{NTU-RGB+D 120~\cite{liu2019ntu}} is an extension of NTU-RGB+D, where the number of classes is expanded to 120 and the number of samples is expanded to 114,480. There are also two recommended evaluation bnechmarks: \textbf{1) Cross-subject (X-sub)} contains 63,026 clips from 53 subjects are used for training, and 50,922 clips from the remaining subjects are reserved for evaluation.\textbf{2) Cross-setup (X-setup)} contains 54,471 clips with even collection setup IDs are used for training, and the rest 59,477 clips with odd setup IDs are used for evaluation. We ignore about 532 bad cases of this dataset according to \cite{liu2019ntu} in our experiments.

\noindent \textbf{Kinetics Skeleton 400~\cite{stgcn2018aaai}} is proposed by \cite{stgcn2018aaai} and the joint location in it is estimated from the Kinetics400 action recognition dataset by using Openpose pose estimation toolbox. It contains 240,436 training and 19,796 evaluation skeleton clips over 400 classes. There are 18 joints in each frames, along with their 2D spatial coordinates and the prediction confidence score from OpenPose as the initial joint features. The evaluation protocol is recommended with Top-1 and Top-5 classification accuracy by the author of the dataset~\cite{carreira2017quo}.

The location of joint annotations in NTU-RGB+D and Kinetics-skeleton can be referred to in Fig.\ref{fig:skeleton}. We heuristically split the human skeleton into three different grains for NTU-RGB+D, and two different grain for Kinetics-skeleton.

\subsection{Implementation Details}

Our codes are implemented based the PyTorch deep learning framework. Stochastic gradient descent (SGD) with Nesterov momentum set 0.9 is applied as the optimization strategy. The batch size is 32. Cross-entropy is selected as the loss function to backpropagate gradients. The weight decay is set to 0.0001.

For NTU-RGB+D datasets, following \cite{2sagcn2019cvpr}, we align the temporal length into 300 frames and repeat the samples to 300 frames when the length is less than 300. The initial learning rate is 0.1 and decays 0.1 at $30_{th}$ and $40_{th}$ epochs with total of 65 epochs. The data pre-processing in \cite{shi2020skeleton} is adopted for NTU-RGB+D and  NTU-RGB+D 120 datasets.

The input size of Kinetics-skeleton is the same as \cite{stgcn2018aaai} with input of 150 frames and 2 bodies in each sample. Also we do the same data augmentation in \cite{stgcn2018aaai}. The learning rate is also set as 0.1 and is decayed 0.1 at the $30_{th}$ epoch and $55_{th}$ epochs with total of 65 epochs.

\subsection{Comparisons with SOTA Methods}

\begin{table}[thb]
\caption{Comparison of the state-of-the-art methods on the NTU RGB+D 120 dataset in terms
of the Top-1 accuracy. $\dagger$ denotes our implemented results based on their released codes on GitHub.}
\centering
\begin{tabular}{lll}
\toprule
Model        & X-Sub & X-Setup \\ \midrule\midrule
ST-GCN~\cite{stgcn2018aaai}       & 70.7$^\dagger$  & 73.2$^\dagger$    \\
RA-GCN~\cite{song2020richly}       & 74.6$^\dagger$  & 75.3$^\dagger$    \\
4s CDGC~\cite{miao2021central}      & 86.3  & 87.8 \\
2s Graph2Net~\cite{wu2021graph2net}	& 86.0	& 87.6 \\
AS-GCN~\cite{Li_2019_CVPR_ASGCN}       & 77.9$^\dagger$  & 78.5$^\dagger$    \\
2s AGCN~\cite{2sagcn2019cvpr}      & 82.5$^\dagger$  & 84.2$^\dagger$    \\
4s Shift-GCN~\cite{DBLP:conf/cvpr/Cheng0HC0L20} & 85.9  & 87.6    \\
SGN~\cite{DBLP:conf/cvpr/0005LZXXZ20}          & 79.2  & 81.5    \\
2s ST-TR~\cite{DBLP:conf/icpr/PlizzariCM20}        & 85.1  & 87.1    \\ \midrule\midrule
4s STF-Net   & 86.5  & 88.2    \\ \bottomrule
\end{tabular}
\label{table:ntu120}
\end{table}

\begin{table}[thb]
\caption{ Comparison of the state-of-the-art methods on the Kinetics Skeleton 400 dataset in terms of the
Top-1 accuracy.}
\centering
\begin{tabular}{lll}
\toprule
Model    & Top1 & Top5 \\ \midrule\midrule
TCN~\cite{DBLP:conf/cvpr/KimR17}      & 20.3 & 40.0 \\
ST-GCN~\cite{stgcn2018aaai}   & 30.7 & 52.8 \\
STGR-GCN~\cite{li2019spatio} & 34.8 & 56.5 \\
AS-GCN~\cite{Li_2019_CVPR_ASGCN}   & 34.8 & 56.5 \\
2s-AGCN~\cite{2sagcn2019cvpr}  & 36.1 & 58.7 \\ 
CA-GCN~\cite{zhang2020context}  & 34.1 & 56.6 \\ \midrule\midrule
4s STF-Net  & 36.1 & 58.9 \\ \bottomrule
\end{tabular}
\label{table:kinetics}
\end{table}

As shown in Table.~\ref{table:ntu60}, our proposed method STF-Net reaches an excellent performance among these SOTA methods, with accuracy 91.1\% for the X-Sub benchmark and 96.5\% for the X-view benchmark. Many current state-of-the-art works follow the multi-stream method, thus for a fair comparison, there are four settings of our STF-Net: \textit{1) Js stream}, which only uses joint modality; \textit{2) Bs stream}, which only uses bone modality; \textit{3) 2s stream}, which uses joint and bone modalities; \textit{4) 4s stream}, which uses joint and bone modalities and their corresponding motion information. 

There are two typical methods we should notice. The first one is STGCN\cite{stgcn2018aaai}, which could be regarded as the pioneering work by applying GCN in the skeleton-based action recognition tasks. It also has been followed by the most current works, our STF-Net is one of them which based on ST-GCN. Compared with ST-GCN in joint modality, our proposed STF-Net outperforms by approximately 7.5\% on the X-sub benchmark and about 6.7\% on X-view benckmark. The second 4s MS-AAGCN, as an extention of 2s-AGCN, is also a popular baseline in current works. We can observe that the accuracy of STF-Net greatly surpasses both benchmarks when using the same modalities and requiring over 2 times fewer parameters. On the NTU-120 RGB+D dataset, as shown in Table.~\ref{table:ntu120}, Our proposed STF-Net obviously exceed previously reported performance, demonstrating the effectiveness of our proposed STF-Net.

On Kinetics, as shown in Table.~\ref{table:kinetics}, we see that our method improves the performance of ST-GCN  by 5.4\% and 6.1\% in Top-1 and Top5 accuracies separately.

\begin{figure}[t]
\centering
\subfigure[brushing teeth]{\includegraphics[width=8cm]{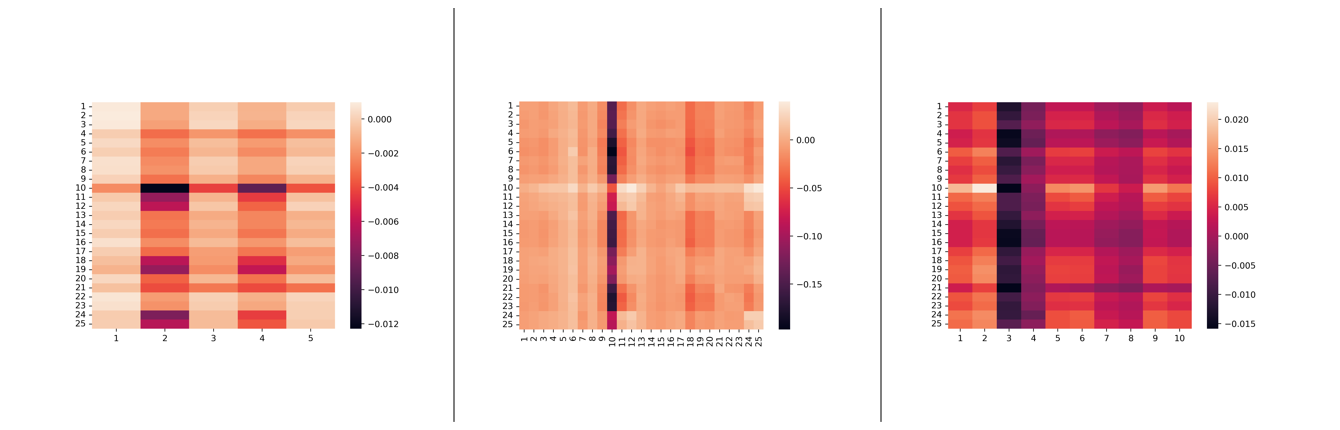}}
\\ %换行
\centering
\subfigure[salute]{\includegraphics[width=8cm]{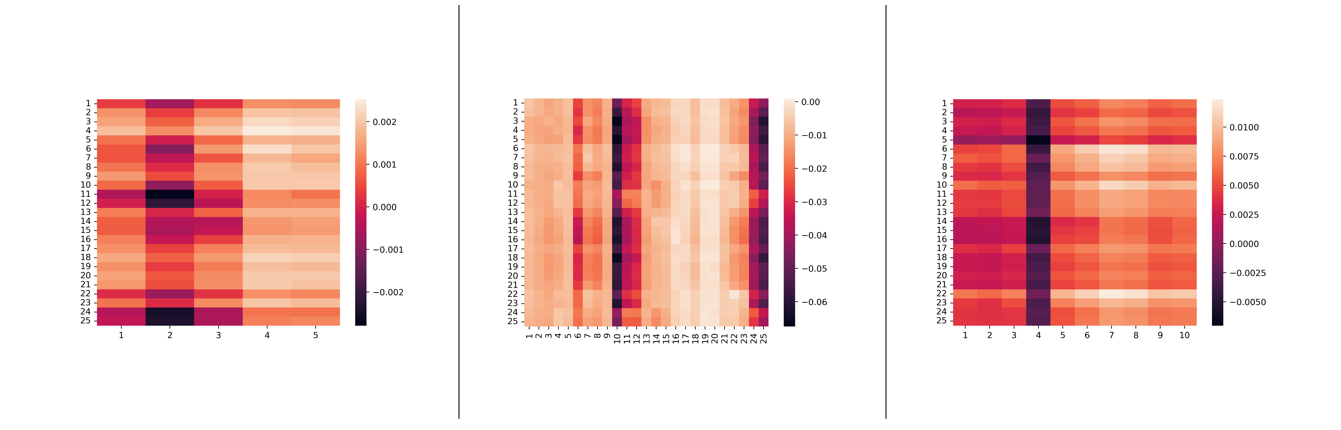}}
\\
\centering
\subfigure[pointing to something with finger]{\includegraphics[width=8cm]{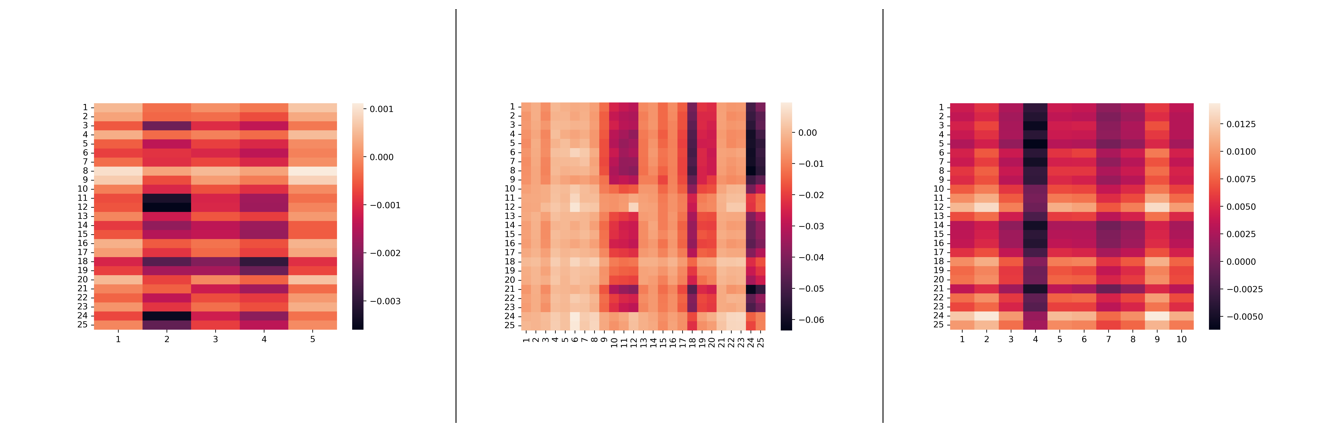}}
\\
\centering
\subfigure[sitting down]{\includegraphics[width=8cm]{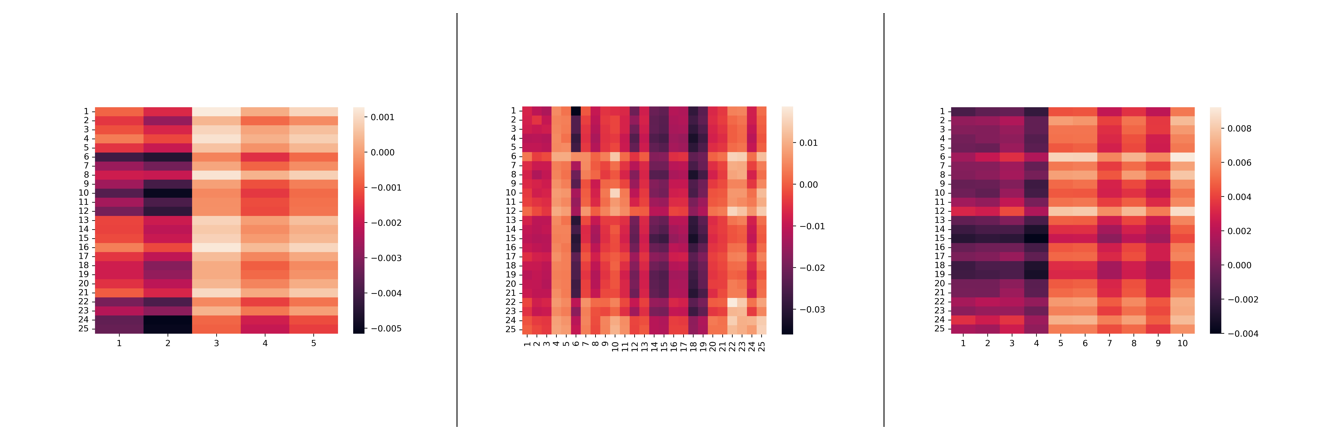}}
\\
\caption{The similarity matrix shows the context among each joint and part in four action instances. The left column represents the relationship in coarse parts and joins. The middle column represents the relationship in joints. The right column represents the relationship in fined part. The brighter of color in the matrix shows the higher strength of connection between two body components (body joint or part).} %图片标题
\label{fig:matrix}
\end{figure}

\begin{figure}[thb]
\centering
\subfigure[drinking water]{\includegraphics[width=8cm]{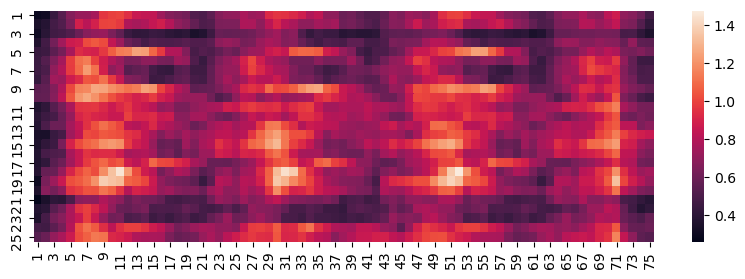}}
\\ %换行
\centering
\subfigure[salute]{\includegraphics[width=8cm]{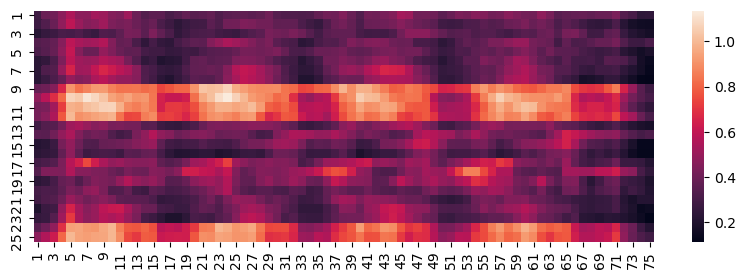}}
\\
\centering
\subfigure[typing on a keyboard]{\includegraphics[width=8cm]{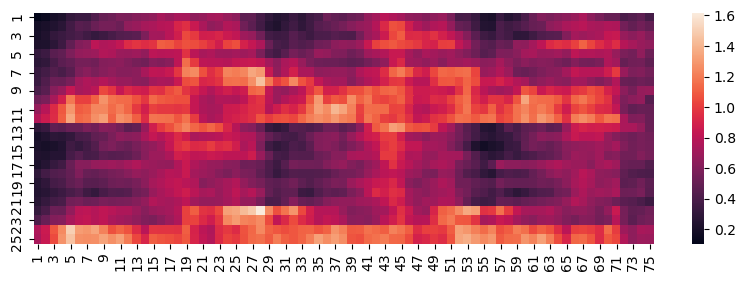}}
\\
\caption{The visualization of the output feature map of STF-Net, where the x-axis shows the temporal dimension and the y-axis shows the spatial dimension. The brighter of color in the feature map shows the crucial duration in the sequence.}

\label{fig:temporal}
\end{figure}

\subsection{Ablation Studies}

\noindent\textbf{Visualization.} To further evaluate out STF-Net, we analyze the learnt context weight among each joint and part. Fig.~\ref{fig:matrix} shows the learned similarity matrix in MCF. The color brightness of the similarity matrix represents the strength of the connection. From left to right three different scales are coarse-part, joint, and fined-part of four action instances respectively. The most interesting finding is that in action brushing teeth, components of left arm take the most important role in this action. In the fined-part of action brushing teeth, we can see that joint 10, elbow, plays the role of the pivot point in fined-part 1 and 2, which represents the left arm has weak connections in the components of the lower body. In contrast, the lower body components show the priority in the action salute. The visualization shows that the weights shown in the similarity matrix are reasonable for those action instances and able to capture the information from the global spatial domain.

Deep analysis in temporal dynamics is shown in Fig~\ref{fig:temporal}. Firstly, the heatmaps depict the output feature map of STF-Net. For instance, in the action drinking water, we could observe different temporal dynamics in each spatial joint, showing that each spatial joint plays an individual role according to the action instance. The heatmap of typing on a keyboard has an obvious contrast around the whole spatiotemporal domain, the joint index around 5 to 12 and 22 to 24 which indicates the arms and hands component plays an essential role in the whole temporal dimension. Additionally, the temporal axis distribution indicates that diversities exist in different action sequences,and it also shows that our model learns the sensitive temporal information indeedly.

\noindent\textbf{Per-class accuracy.} Per-class accuracy increment can be found in Fig~\ref{fig:acc}, where all categories are listed in the x-axis. The overall accuracy of each class has been promoted compared with baseline ST-GCN. Besides, actions like reading, playing with phone/tablet, and sneeze/cough that get the long distance between two components in physical structure have achieved the higher improvements. We achieve an almost 2.0\% relative gain in accuracy compared to the baseline. For simplicity, the comparison in Fig.~\ref{fig:acc} is conducted on the NTU-TGB+D 60 dataset x-sub benchmark with joint modality.

% \usepackage{multirow}
\begin{table}[thb]
\caption{Ablation study on different components of STF-Net. $\dagger$ denotes our implemented results. w/o denotes the model without the module.}
\centering
\begin{tabular}{lllll}
\toprule
Model                    & Dataset                   & \#Parts            & Top1 & Top5 \\ \midrule\midrule
\multirow{5}{*}{STF-Net} & \multirow{2}{*}{Kinetics} & 1                  & 34.4 & 57.1 \\
                         &                           & 2                  & 35.0 & 57.6 \\ \cline{2-5} 
                         & \multirow{3}{*}{NTU60}    & 1                  & 88.2 & 97.6 \\
                         &                           & 2                  & 88.4 & 97.9 \\
                         &                           & 3                  & 88.8 & 97.9 \\ \midrule\midrule
ST-GCN$^\dagger$                    & \multirow{4}{*}{NTU60}    & \multirow{3}{*}{-} & 86.2 & 96.5 \\
STF-Net w/o MCF\&TDF   &                           &                    & 87.4 & 97.2 \\
STF-Net w/o MCF         &                           &                    & 88.1 & 97.9 \\
STF-Net w/o TDF         &                           & 3                  & 88.0 & 97.7 \\ \bottomrule
\end{tabular}
\label{table:module}
\end{table}

\noindent\textbf{Ablation on module decoupling in STF-Net.} As shown in Table.~\ref{table:module}, with the module decoupling in STF-Net. The table could be interpreted into two aspects. the part above the double horizontal lines of the table shows that the number of human parts which we split affects the performance of the model. We conduct the ablation experiments on the Kinetics and NTU60 datasets respectively. As the number of the parts increases, the performance has a gradual improvement in 0.2\% and 0.6\% in NTU60, showing that different scales of parts contains complementary information of skeleton recognition. The bottom of the table shows the ablation of components of MCF and TDF. For fair comparison, we re-imply our backbone ST-GCN, the Top1 and Top5 accuracies are 86.2\% and 96.5\%. While adding graph mask into ST-GCN, STF-Net w/o MCF\&TDF in the table, the Top1 accuracy has improved about 1.2\%. The following ablation studies are on module MCF and TDF, when only TDF is adopted, the Top1 accuracy reaches 88.1\%, which surpasses the baseline by over 2\%. In addition, STF-Net with only MCF can also achieve over 88.0\%.

\begin{figure}[thb]
   \centering
   \includegraphics[width=8cm]{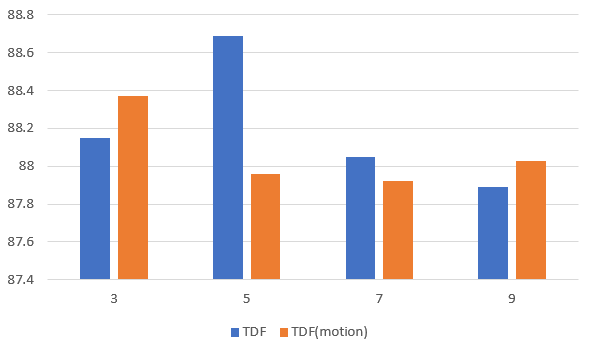}
   \caption{Results of models with different kernel size in TDF.}
\label{fig:TDF}
\end{figure}

\noindent\textbf{Ablation on kernel size of TDF.} TDF is designed for capturing the discriminated temporal dynamics. Here we set two different designed TDF by adding motion difference. In the experiments, we set the kernel size of TCN as 9 according to the best option in \cite{zhang2020context}.  Fig.~\ref{fig:TDF} shows  the Top-1 accuracy of our model with different types of TDF, we can see that the models perform worse in both TDF and TDF(motion) if when the kernel size is getting larger. Thus, we induce that include too many frames in TDF may cause distant frames to learn the discrimination of short-term temporal dynamics around the current frame. In contrast, when we chose kernel size as 5 in TDF and kernel size as 3 in TDF(motion), the models could reach the optimal performance when we chose kernel size as 5 in TDF and kernel size as 3 in TDF(motion).

\section{Conclusion}
In this paper, we propose a multi-grain contextual focus module (MCF) and a temporal discrimination focus module (TDF) based on ST-GCN to form our designed spatiotemporal focus network (STF-Net) for skeleton-based action recognition. Inspired by the semantic contained in different scales of parts, we heuristically split several scales in the skeleton and use self-attention methods to capture the relations among them. Besides, we try to enhance temporal dynamics due to the dense sampling strategy and the redundancy among each frame in the skeleton sequence by the method of squeeze and excitation.

Experiments on three large-scale benchmarks show that our model outperforms existing methods by a sizable margin. Additionally, the visualization results verify that STF-Net captures the contextual information in spatial dimension and the discriminated dynamics in the temporal dimension depending on different action instances.

% or


\appendices

% % you can choose not to have a title for an appendix
% % if you want by leaving the argument blank
% \section{}

% % use section* for acknowledgment
% \section*{Acknowledgment}


\ifCLASSOPTIONcaptionsoff
  \newpage
\fi

\bibliographystyle{IEEEtran}

\begin{thebibliography}{10}
\providecommand{\url}[1]{#1}
\csname url@samestyle\endcsname
\providecommand{\newblock}{\relax}
\providecommand{\bibinfo}[2]{#2}
\providecommand{\BIBentrySTDinterwordspacing}{\spaceskip=0pt\relax}
\providecommand{\BIBentryALTinterwordstretchfactor}{4}
\providecommand{\BIBentryALTinterwordspacing}{\spaceskip=\fontdimen2\font plus
\BIBentryALTinterwordstretchfactor\fontdimen3\font minus
  \fontdimen4\font\relax}
\providecommand{\BIBforeignlanguage}[2]{{%
\expandafter\ifx\csname l@#1\endcsname\relax
\typeout{** WARNING: IEEEtran.bst: No hyphenation pattern has been}%
\typeout{** loaded for the language `#1'. Using the pattern for}%
\typeout{** the default language instead.}%
\else
\language=\csname l@#1\endcsname
\fi
#2}}
\providecommand{\BIBdecl}{\relax}
\BIBdecl

\bibitem{in_GraphNet_ijcai2020}
D.~Yang and Y.~Zou, ``A graph-based interactive reasoning for human-object
  interaction detection,'' in \emph{Proceedings of the Twenty-Ninth
  International Joint Conference on Artificial Intelligence, {IJCAI-20}}.\hskip
  1em plus 0.5em minus 0.4em\relax International Joint Conferences on
  Artificial Intelligence Organization, 2020, pp. 1111--1117.

\bibitem{yang2021rr_net}
D.~Yang, Y.~Zou, C.~Zhang, M.~Cao, and J.~Chen, ``Rr-net: Relation reasoning
  for end-to-end human-object interaction detection,'' \emph{IEEE Transactions
  on Circuits and Systems for Video Technology}, 2021.

\bibitem{Zhu_Lan_Xing_Zeng_Li_Shen_Xie_2016_depth_sensor}
\BIBentryALTinterwordspacing
W.~Zhu, C.~Lan, J.~Xing, W.~Zeng, Y.~Li, L.~Shen, and X.~Xie, ``Co-occurrence
  feature learning for skeleton based action recognition using regularized deep
  lstm networks,'' \emph{Proceedings of the AAAI Conference on Artificial
  Intelligence}, vol.~30, no.~1, Mar. 2016. [Online]. Available:
  \url{https://ojs.aaai.org/index.php/AAAI/article/view/10451}
\BIBentrySTDinterwordspacing

\bibitem{Cao_2017_CVPR_pose_estimation}
Z.~Cao, T.~Simon, S.-E. Wei, and Y.~Sheikh, ``Realtime multi-person 2d pose
  estimation using part affinity fields,'' in \emph{Proceedings of the IEEE
  Conference on Computer Vision and Pattern Recognition (CVPR)}, July 2017.

\bibitem{Sun_2019_CVPR_pose_estimation}
K.~Sun, B.~Xiao, D.~Liu, and J.~Wang, ``Deep high-resolution representation
  learning for human pose estimation,'' in \emph{Proceedings of the IEEE/CVF
  Conference on Computer Vision and Pattern Recognition (CVPR)}, June 2019.

\bibitem{openpose}
\BIBentryALTinterwordspacing
Z.~Cao, G.~Hidalgo, T.~Simon, S.~Wei, and Y.~Sheikh, ``Openpose: Realtime
  multi-person 2d pose estimation using part affinity fields,'' \emph{CoRR},
  vol. abs/1812.08008, 2018. [Online]. Available:
  \url{http://arxiv.org/abs/1812.08008}
\BIBentrySTDinterwordspacing

\bibitem{skeleton_modality}
S.~{Song}, C.~{Lan}, J.~{Xing}, W.~{Zeng}, and J.~{Liu}, ``Skeleton-indexed
  deep multi-modal feature learning for high performance human action
  recognition,'' in \emph{2018 IEEE International Conference on Multimedia and
  Expo (ICME)}, 2018, pp. 1--6.

\bibitem{Ke_2017_CVPR_RNN}
Q.~Ke, M.~Bennamoun, S.~An, F.~Sohel, and F.~Boussaid, ``A new representation
  of skeleton sequences for 3d action recognition,'' in \emph{Proceedings of
  the IEEE Conference on Computer Vision and Pattern Recognition (CVPR)}, July
  2017.

\bibitem{Kim_2017_CVPR_Workshops_RNN}
T.~Soo~Kim and A.~Reiter, ``Interpretable 3d human action analysis with
  temporal convolutional networks,'' in \emph{Proceedings of the IEEE
  Conference on Computer Vision and Pattern Recognition (CVPR) Workshops}, July
  2017.

\bibitem{Du_2015_CVPR_RNN}
Y.~Du, W.~Wang, and L.~Wang, ``Hierarchical recurrent neural network for
  skeleton based action recognition,'' in \emph{Proceedings of the IEEE
  Conference on Computer Vision and Pattern Recognition (CVPR)}, June 2015.

\bibitem{ST_LSTM_RNN}
J.~Liu, A.~Shahroudy, D.~Xu, and G.~Wang, ``Spatio-temporal lstm with trust
  gates for 3d human action recognition,'' in \emph{Computer Vision -- ECCV
  2016}, B.~Leibe, J.~Matas, N.~Sebe, and M.~Welling, Eds.\hskip 1em plus 0.5em
  minus 0.4em\relax Cham: Springer International Publishing, 2016, pp.
  816--833.

\bibitem{ICMEW_2017_CNN}
{Chao Li}, {Qiaoyong Zhong}, {Di Xie}, and {Shiliang Pu}, ``Skeleton-based
  action recognition with convolutional neural networks,'' in \emph{2017 IEEE
  International Conference on Multimedia Expo Workshops (ICMEW)}, 2017, pp.
  597--600.

\bibitem{IJCAI_2018_CNN}
\BIBentryALTinterwordspacing
C.~Li, Q.~Zhong, D.~Xie, and S.~Pu, ``Co-occurrence feature learning from
  skeleton data for action recognition and detection with hierarchical
  aggregation,'' \emph{CoRR}, vol. abs/1804.06055, 2018. [Online]. Available:
  \url{http://arxiv.org/abs/1804.06055}
\BIBentrySTDinterwordspacing

\bibitem{Goated_CNN}
C.~{Cao}, C.~{Lan}, Y.~{Zhang}, W.~{Zeng}, H.~{Lu}, and Y.~{Zhang},
  ``Skeleton-based action recognition with gated convolutional neural
  networks,'' \emph{IEEE Transactions on Circuits and Systems for Video
  Technology}, vol.~29, no.~11, pp. 3247--3257, 2019.

\bibitem{stgcn2018aaai}
\BIBentryALTinterwordspacing
S.~Yan, Y.~Xiong, and D.~Lin, ``Spatial temporal graph convolutional networks
  for skeleton-based action recognition,'' in \emph{Proceedings of the
  Thirty-Second {AAAI} Conference on Artificial Intelligence, (AAAI-18), the
  30th innovative Applications of Artificial Intelligence (IAAI-18), and the
  8th {AAAI} Symposium on Educational Advances in Artificial Intelligence
  (EAAI-18), New Orleans, Louisiana, USA, February 2-7, 2018}, S.~A. McIlraith
  and K.~Q. Weinberger, Eds.\hskip 1em plus 0.5em minus 0.4em\relax {AAAI}
  Press, 2018, pp. 7444--7452. [Online]. Available:
  \url{https://www.aaai.org/ocs/index.php/AAAI/AAAI18/paper/view/17135}
\BIBentrySTDinterwordspacing

\bibitem{2sagcn2019cvpr}
\BIBentryALTinterwordspacing
L.~Shi, Y.~Zhang, J.~Cheng, and H.~Lu, ``Two-stream adaptive graph
  convolutional networks for skeleton-based action recognition,'' in
  \emph{{IEEE} Conference on Computer Vision and Pattern Recognition, {CVPR}
  2019, Long Beach, CA, USA, June 16-20, 2019}.\hskip 1em plus 0.5em minus
  0.4em\relax Computer Vision Foundation / {IEEE}, 2019, pp. 12\,026--12\,035.
  [Online]. Available:
  \url{http://openaccess.thecvf.com/content\_CVPR\_2019/html/Shi\_Two-Stream\_Adaptive\_Graph\_Convolutional\_Networks\_for\_Skeleton-Based\_Action\_Recognition\_CVPR\_2019\_paper.html}
\BIBentrySTDinterwordspacing

\bibitem{Li_2019_CVPR_ASGCN}
M.~Li, S.~Chen, X.~Chen, Y.~Zhang, Y.~Wang, and Q.~Tian, ``Actional-structural
  graph convolutional networks for skeleton-based action recognition,'' in
  \emph{Proceedings of the IEEE/CVF Conference on Computer Vision and Pattern
  Recognition (CVPR)}, June 2019.

\bibitem{shi2019skeleton}
\BIBentryALTinterwordspacing
L.~Shi, Y.~Zhang, J.~Cheng, and H.~Lu, ``Skeleton-based action recognition with
  directed graph neural networks,'' in \emph{{IEEE} Conference on Computer
  Vision and Pattern Recognition, {CVPR} 2019, Long Beach, CA, USA, June 16-20,
  2019}.\hskip 1em plus 0.5em minus 0.4em\relax Computer Vision Foundation /
  {IEEE}, 2019, pp. 7912--7921. [Online]. Available:
  \url{http://openaccess.thecvf.com/content\_CVPR\_2019/html/Shi\_Skeleton-Based\_Action\_Recognition\_With\_Directed\_Graph\_Neural\_Networks\_CVPR\_2019\_paper.html}
\BIBentrySTDinterwordspacing

\bibitem{Liu_2020_CVPR_G3D}
Z.~Liu, H.~Zhang, Z.~Chen, Z.~Wang, and W.~Ouyang, ``Disentangling and unifying
  graph convolutions for skeleton-based action recognition,'' in
  \emph{Proceedings of the IEEE/CVF Conference on Computer Vision and Pattern
  Recognition (CVPR)}, June 2020.

\bibitem{zhang2020context}
X.~Zhang, C.~Xu, and D.~Tao, ``Context aware graph convolution for
  skeleton-based action recognition,'' in \emph{Proceedings of the IEEE/CVF
  Conference on Computer Vision and Pattern Recognition}, 2020, pp.
  14\,333--14\,342.

\bibitem{song2020richly}
Y.-F. Song, Z.~Zhang, C.~Shan, and L.~Wang, ``Richly activated graph
  convolutional network for robust skeleton-based action recognition,''
  \emph{IEEE Transactions on Circuits and Systems for Video Technology},
  vol.~31, no.~5, pp. 1915--1925, 2020.

\bibitem{yang2018action}
Z.~Yang, Y.~Li, J.~Yang, and J.~Luo, ``Action recognition with spatio--temporal
  visual attention on skeleton image sequences,'' \emph{IEEE Transactions on
  Circuits and Systems for Video Technology}, vol.~29, no.~8, pp. 2405--2415,
  2018.

\bibitem{wang2021multi}
H.~Wang, B.~Yu, J.~Li, L.~Zhang, and D.~Chen, ``Multi-stream interaction
  networks for human action recognition,'' \emph{IEEE Transactions on Circuits
  and Systems for Video Technology}, 2021.

\bibitem{miao2021central}
S.~Miao, Y.~Hou, Z.~Gao, M.~Xu, and W.~Li, ``A central difference graph
  convolutional operator for skeleton-based action recognition,'' \emph{IEEE
  Transactions on Circuits and Systems for Video Technology}, 2021.

\bibitem{wu2021graph2net}
C.~Wu, X.-J. Wu, and J.~Kittler, ``Graph2net: Perceptually-enriched graph
  learning for skeleton-based action recognition,'' \emph{IEEE Transactions on
  Circuits and Systems for Video Technology}, 2021.

\bibitem{DBLP:conf/eccv/LinZSWY18}
\BIBentryALTinterwordspacing
T.~Lin, X.~Zhao, H.~Su, C.~Wang, and M.~Yang, ``{BSN:} boundary sensitive
  network for temporal action proposal generation,'' in \emph{Computer Vision -
  {ECCV} 2018 - 15th European Conference, Munich, Germany, September 8-14,
  2018, Proceedings, Part {IV}}, ser. Lecture Notes in Computer Science,
  V.~Ferrari, M.~Hebert, C.~Sminchisescu, and Y.~Weiss, Eds., vol. 11208.\hskip
  1em plus 0.5em minus 0.4em\relax Springer, 2018, pp. 3--21. [Online].
  Available: \url{https://doi.org/10.1007/978-3-030-01225-0\_1}
\BIBentrySTDinterwordspacing

\bibitem{zhu2019asymmetric}
Z.~Zhu, M.~Xu, S.~Bai, T.~Huang, and X.~Bai, ``Asymmetric non-local neural
  networks for semantic segmentation,'' in \emph{Proceedings of the IEEE/CVF
  International Conference on Computer Vision}, 2019, pp. 593--602.

\bibitem{hu2018squeeze}
J.~Hu, L.~Shen, and G.~Sun, ``Squeeze-and-excitation networks,'' in
  \emph{Proceedings of the IEEE conference on computer vision and pattern
  recognition}, 2018, pp. 7132--7141.

\bibitem{le2011learning}
Q.~V. Le, W.~Y. Zou, S.~Y. Yeung, and A.~Y. Ng, ``Learning hierarchical
  invariant spatio-temporal features for action recognition with independent
  subspace analysis,'' in \emph{CVPR 2011}.\hskip 1em plus 0.5em minus
  0.4em\relax IEEE, 2011, pp. 3361--3368.

\bibitem{sadanand2012action}
S.~Sadanand and J.~J. Corso, ``Action bank: A high-level representation of
  activity in video,'' in \emph{2012 IEEE Conference on Computer Vision and
  Pattern Recognition}.\hskip 1em plus 0.5em minus 0.4em\relax IEEE, 2012, pp.
  1234--1241.

\bibitem{wang2018temporal}
L.~Wang, Y.~Xiong, Z.~Wang, Y.~Qiao, D.~Lin, X.~Tang, and L.~Van~Gool,
  ``Temporal segment networks for action recognition in videos,'' \emph{IEEE
  transactions on pattern analysis and machine intelligence}, vol.~41, no.~11,
  pp. 2740--2755, 2018.

\bibitem{lin2019tsm}
J.~Lin, C.~Gan, and S.~Han, ``Tsm: Temporal shift module for efficient video
  understanding,'' in \emph{Proceedings of the IEEE/CVF International
  Conference on Computer Vision}, 2019, pp. 7083--7093.

\bibitem{tran2015learning}
D.~Tran, L.~Bourdev, R.~Fergus, L.~Torresani, and M.~Paluri, ``Learning
  spatiotemporal features with 3d convolutional networks,'' in
  \emph{Proceedings of the IEEE international conference on computer vision},
  2015, pp. 4489--4497.

\bibitem{carreira2017quo}
J.~Carreira and A.~Zisserman, ``Quo vadis, action recognition? a new model and
  the kinetics dataset,'' in \emph{proceedings of the IEEE Conference on
  Computer Vision and Pattern Recognition}, 2017, pp. 6299--6308.

\bibitem{Wang_2018_CVPR_nonlocal}
X.~Wang, R.~Girshick, A.~Gupta, and K.~He, ``Non-local neural networks,'' in
  \emph{Proceedings of the IEEE Conference on Computer Vision and Pattern
  Recognition (CVPR)}, June 2018.

\bibitem{sun2018optical}
S.~Sun, Z.~Kuang, L.~Sheng, W.~Ouyang, and W.~Zhang, ``Optical flow guided
  feature: A fast and robust motion representation for video action
  recognition,'' in \emph{Proceedings of the IEEE conference on computer vision
  and pattern recognition}, 2018, pp. 1390--1399.

\bibitem{zhang2019pan}
C.~Zhang, Y.~Zou, G.~Chen, and L.~Gan, ``Pan: Persistent appearance network
  with an efficient motion cue for fast action recognition,'' in
  \emph{Proceedings of the 27th ACM International Conference on Multimedia},
  2019, pp. 500--509.

\bibitem{DBLP:conf/cvpr/DuWW15}
\BIBentryALTinterwordspacing
Y.~Du, W.~Wang, and L.~Wang, ``Hierarchical recurrent neural network for
  skeleton based action recognition,'' in \emph{{IEEE} Conference on Computer
  Vision and Pattern Recognition, {CVPR} 2015, Boston, MA, USA, June 7-12,
  2015}.\hskip 1em plus 0.5em minus 0.4em\relax {IEEE} Computer Society, 2015,
  pp. 1110--1118. [Online]. Available:
  \url{https://doi.org/10.1109/CVPR.2015.7298714}
\BIBentrySTDinterwordspacing

\bibitem{shahroudy2016ntu}
A.~Shahroudy, J.~Liu, T.-T. Ng, and G.~Wang, ``Ntu rgb+ d: A large scale
  dataset for 3d human activity analysis,'' in \emph{Proceedings of the IEEE
  conference on computer vision and pattern recognition}, 2016, pp. 1010--1019.

\bibitem{DBLP:conf/avss/CaetanoSBSS19}
\BIBentryALTinterwordspacing
C.~Caetano, J.~Sena, F.~Br{\'{e}}mond, J.~A. dos Santos, and W.~R. Schwartz,
  ``Skelemotion: {A} new representation of skeleton joint sequences based on
  motion information for 3d action recognition,'' in \emph{16th {IEEE}
  International Conference on Advanced Video and Signal Based Surveillance,
  {AVSS} 2019, Taipei, Taiwan, September 18-21, 2019}.\hskip 1em plus 0.5em
  minus 0.4em\relax {IEEE}, 2019, pp. 1--8. [Online]. Available:
  \url{https://doi.org/10.1109/AVSS.2019.8909840}
\BIBentrySTDinterwordspacing

\bibitem{DBLP:conf/icpr/PlizzariCM20}
\BIBentryALTinterwordspacing
C.~Plizzari, M.~Cannici, and M.~Matteucci, ``Spatial temporal transformer
  network for skeleton-based action recognition,'' in \emph{Pattern
  Recognition. {ICPR} International Workshops and Challenges - Virtual Event,
  January 10-15, 2021, Proceedings, Part {III}}, ser. Lecture Notes in Computer
  Science, A.~D. Bimbo, R.~Cucchiara, S.~Sclaroff, G.~M. Farinella, T.~Mei,
  M.~Bertini, H.~J. Escalante, and R.~Vezzani, Eds., vol. 12663.\hskip 1em plus
  0.5em minus 0.4em\relax Springer, 2020, pp. 694--701. [Online]. Available:
  \url{https://doi.org/10.1007/978-3-030-68796-0\_50}
\BIBentrySTDinterwordspacing

\bibitem{bruna2014spectral}
\BIBentryALTinterwordspacing
J.~Bruna, W.~Zaremba, A.~Szlam, and Y.~LeCun, ``Spectral networks and locally
  connected networks on graphs,'' in \emph{2nd International Conference on
  Learning Representations, {ICLR} 2014, Banff, AB, Canada, April 14-16, 2014,
  Conference Track Proceedings}, Y.~Bengio and Y.~LeCun, Eds., 2014. [Online].
  Available: \url{http://arxiv.org/abs/1312.6203}
\BIBentrySTDinterwordspacing

\bibitem{defferrard2017convolutional_spectral}
\BIBentryALTinterwordspacing
M.~Defferrard, X.~Bresson, and P.~Vandergheynst, ``Convolutional neural
  networks on graphs with fast localized spectral filtering,'' in
  \emph{Advances in Neural Information Processing Systems 29: Annual Conference
  on Neural Information Processing Systems 2016, December 5-10, 2016,
  Barcelona, Spain}, D.~D. Lee, M.~Sugiyama, U.~von Luxburg, I.~Guyon, and
  R.~Garnett, Eds., 2016, pp. 3837--3845. [Online]. Available:
  \url{https://proceedings.neurips.cc/paper/2016/hash/04df4d434d481c5bb723be1b6df1ee65-Abstract.html}
\BIBentrySTDinterwordspacing

\bibitem{henaff2015deep_spectral}
\BIBentryALTinterwordspacing
M.~Henaff, J.~Bruna, and Y.~LeCun, ``Deep convolutional networks on
  graph-structured data,'' \emph{CoRR}, vol. abs/1506.05163, 2015. [Online].
  Available: \url{http://arxiv.org/abs/1506.05163}
\BIBentrySTDinterwordspacing

\bibitem{Shuman_2013_spectral}
\BIBentryALTinterwordspacing
D.~I. Shuman, S.~K. Narang, P.~Frossard, A.~Ortega, and P.~Vandergheynst, ``The
  emerging field of signal processing on graphs: Extending high-dimensional
  data analysis to networks and other irregular domains,'' \emph{IEEE Signal
  Processing Magazine}, vol.~30, no.~3, p. 83–98, May 2013. [Online].
  Available: \url{http://dx.doi.org/10.1109/MSP.2012.2235192}
\BIBentrySTDinterwordspacing

\bibitem{niepert2016learning_spatail}
\BIBentryALTinterwordspacing
M.~Niepert, M.~Ahmed, and K.~Kutzkov, ``Learning convolutional neural networks
  for graphs,'' in \emph{Proceedings of the 33nd International Conference on
  Machine Learning, {ICML} 2016, New York City, NY, USA, June 19-24, 2016},
  ser. {JMLR} Workshop and Conference Proceedings, M.~Balcan and K.~Q.
  Weinberger, Eds., vol.~48.\hskip 1em plus 0.5em minus 0.4em\relax JMLR.org,
  2016, pp. 2014--2023. [Online]. Available:
  \url{http://proceedings.mlr.press/v48/niepert16.html}
\BIBentrySTDinterwordspacing

\bibitem{lin2017feature_pyramid}
T.-Y. Lin, P.~Doll{\'a}r, R.~Girshick, K.~He, B.~Hariharan, and S.~Belongie,
  ``Feature pyramid networks for object detection,'' in \emph{Proceedings of
  the IEEE conference on computer vision and pattern recognition}, 2017, pp.
  2117--2125.

\bibitem{cai2016unified_fast_detection}
Z.~Cai, Q.~Fan, R.~S. Feris, and N.~Vasconcelos, ``A unified multi-scale deep
  convolutional neural network for fast object detection,'' in \emph{European
  conference on computer vision}.\hskip 1em plus 0.5em minus 0.4em\relax
  Springer, 2016, pp. 354--370.

\bibitem{chen2018biglittle}
C.-F. Chen, Q.~Fan, N.~Mallinar, T.~Sercu, and R.~Feris, ``Big-little net: An
  efficient multi-scale feature representation for visual and speech
  recognition,'' \emph{arXiv preprint arXiv:1807.03848}, 2018.

\bibitem{newell2016stacked_hourglass}
A.~Newell, K.~Yang, and J.~Deng, ``Stacked hourglass networks for human pose
  estimation,'' in \emph{European conference on computer vision}.\hskip 1em
  plus 0.5em minus 0.4em\relax Springer, 2016, pp. 483--499.

\bibitem{chen2021crossvit}
C.-F. Chen, Q.~Fan, and R.~Panda, ``Crossvit: Cross-attention multi-scale
  vision transformer for image classification,'' \emph{arXiv preprint
  arXiv:2103.14899}, 2021.

\bibitem{fan2019blvnet}
Q.~Fan, C.-F. Chen, H.~Kuehne, M.~Pistoia, and D.~Cox, ``More is less: Learning
  efficient video representations by big-little network and depthwise temporal
  aggregation,'' \emph{arXiv preprint arXiv:1912.00869}, 2019.

\bibitem{feichtenhofer2019slowfast}
C.~Feichtenhofer, H.~Fan, J.~Malik, and K.~He, ``Slowfast networks for video
  recognition,'' in \emph{Proceedings of the IEEE/CVF International Conference
  on Computer Vision}, 2019, pp. 6202--6211.

\bibitem{DBLP:conf/iclr/LiaoZUZ19}
\BIBentryALTinterwordspacing
R.~Liao, Z.~Zhao, R.~Urtasun, and R.~S. Zemel, ``Lanczosnet: Multi-scale deep
  graph convolutional networks,'' in \emph{7th International Conference on
  Learning Representations, {ICLR} 2019, New Orleans, LA, USA, May 6-9,
  2019}.\hskip 1em plus 0.5em minus 0.4em\relax OpenReview.net, 2019. [Online].
  Available: \url{https://openreview.net/forum?id=BkedznAqKQ}
\BIBentrySTDinterwordspacing

\bibitem{vaswani2017attention}
A.~Vaswani, N.~Shazeer, N.~Parmar, J.~Uszkoreit, L.~Jones, A.~N. Gomez,
  L.~Kaiser, and I.~Polosukhin, ``Attention is all you need,'' \emph{arXiv
  preprint arXiv:1706.03762}, 2017.

\bibitem{huang2020spatio}
Z.~Huang, X.~Shen, X.~Tian, H.~Li, J.~Huang, and X.-S. Hua, ``Spatio-temporal
  inception graph convolutional networks for skeleton-based action
  recognition,'' in \emph{Proceedings of the 28th ACM International Conference
  on Multimedia}, 2020, pp. 2122--2130.

\bibitem{xie2017aggregated}
S.~Xie, R.~Girshick, P.~Doll{\'a}r, Z.~Tu, and K.~He, ``Aggregated residual
  transformations for deep neural networks,'' in \emph{Proceedings of the IEEE
  conference on computer vision and pattern recognition}, 2017, pp. 1492--1500.

\bibitem{shi2020skeleton}
L.~Shi, Y.~Zhang, J.~Cheng, and H.~Lu, ``Skeleton-based action recognition with
  multi-stream adaptive graph convolutional networks,'' \emph{IEEE Transactions
  on Image Processing}, vol.~29, pp. 9532--9545, 2020.

\bibitem{DBLP:conf/cvpr/KimR17}
\BIBentryALTinterwordspacing
T.~S. Kim and A.~Reiter, ``Interpretable 3d human action analysis with temporal
  convolutional networks,'' in \emph{2017 {IEEE} Conference on Computer Vision
  and Pattern Recognition Workshops, {CVPR} Workshops 2017, Honolulu, HI, USA,
  July 21-26, 2017}.\hskip 1em plus 0.5em minus 0.4em\relax {IEEE} Computer
  Society, 2017, pp. 1623--1631. [Online]. Available:
  \url{https://doi.org/10.1109/CVPRW.2017.207}
\BIBentrySTDinterwordspacing

\bibitem{DBLP:journals/pami/ZhangLXZXZ19}
\BIBentryALTinterwordspacing
P.~Zhang, C.~Lan, J.~Xing, W.~Zeng, J.~Xue, and N.~Zheng, ``View adaptive
  neural networks for high performance skeleton-based human action
  recognition,'' \emph{{IEEE} Trans. Pattern Anal. Mach. Intell.}, vol.~41,
  no.~8, pp. 1963--1978, 2019. [Online]. Available:
  \url{https://doi.org/10.1109/TPAMI.2019.2896631}
\BIBentrySTDinterwordspacing

\bibitem{DBLP:conf/aaai/YanXL18}
\BIBentryALTinterwordspacing
S.~Yan, Y.~Xiong, and D.~Lin, ``Spatial temporal graph convolutional networks
  for skeleton-based action recognition,'' in \emph{Proceedings of the
  Thirty-Second {AAAI} Conference on Artificial Intelligence, (AAAI-18), the
  30th innovative Applications of Artificial Intelligence (IAAI-18), and the
  8th {AAAI} Symposium on Educational Advances in Artificial Intelligence
  (EAAI-18), New Orleans, Louisiana, USA, February 2-7, 2018}, S.~A. McIlraith
  and K.~Q. Weinberger, Eds.\hskip 1em plus 0.5em minus 0.4em\relax {AAAI}
  Press, 2018, pp. 7444--7452. [Online]. Available:
  \url{https://www.aaai.org/ocs/index.php/AAAI/AAAI18/paper/view/17135}
\BIBentrySTDinterwordspacing

\bibitem{DBLP:conf/mm/Gao0T0G19}
\BIBentryALTinterwordspacing
X.~Gao, W.~Hu, J.~Tang, J.~Liu, and Z.~Guo, ``Optimized skeleton-based action
  recognition via sparsified graph regression,'' in \emph{Proceedings of the
  27th {ACM} International Conference on Multimedia, {MM} 2019, Nice, France,
  October 21-25, 2019}, L.~Amsaleg, B.~Huet, M.~A. Larson, G.~Gravier, H.~Hung,
  C.~Ngo, and W.~T. Ooi, Eds.\hskip 1em plus 0.5em minus 0.4em\relax {ACM},
  2019, pp. 601--610. [Online]. Available:
  \url{https://doi.org/10.1145/3343031.3351170}
\BIBentrySTDinterwordspacing

\bibitem{DBLP:conf/cvpr/SiC0WT19}
\BIBentryALTinterwordspacing
C.~Si, W.~Chen, W.~Wang, L.~Wang, and T.~Tan, ``An attention enhanced graph
  convolutional {LSTM} network for skeleton-based action recognition,'' in
  \emph{{IEEE} Conference on Computer Vision and Pattern Recognition, {CVPR}
  2019, Long Beach, CA, USA, June 16-20, 2019}.\hskip 1em plus 0.5em minus
  0.4em\relax Computer Vision Foundation / {IEEE}, 2019, pp. 1227--1236.
  [Online]. Available:
  \url{http://openaccess.thecvf.com/content\_CVPR\_2019/html/Si\_An\_Attention\_Enhanced\_Graph\_Convolutional\_LSTM\_Network\_for\_Skeleton-Based\_Action\_CVPR\_2019\_paper.html}
\BIBentrySTDinterwordspacing

\bibitem{DBLP:conf/aaai/HuangHOW20}
\BIBentryALTinterwordspacing
L.~Huang, Y.~Huang, W.~Ouyang, and L.~Wang, ``Part-level graph convolutional
  network for skeleton-based action recognition,'' in \emph{The Thirty-Fourth
  {AAAI} Conference on Artificial Intelligence, {AAAI} 2020, The Thirty-Second
  Innovative Applications of Artificial Intelligence Conference, {IAAI} 2020,
  The Tenth {AAAI} Symposium on Educational Advances in Artificial
  Intelligence, {EAAI} 2020, New York, NY, USA, February 7-12, 2020}.\hskip 1em
  plus 0.5em minus 0.4em\relax {AAAI} Press, 2020, pp. 11\,045--11\,052.
  [Online]. Available:
  \url{https://aaai.org/ojs/index.php/AAAI/article/view/6759}
\BIBentrySTDinterwordspacing

\bibitem{DBLP:conf/aaai/PengHCZ20}
\BIBentryALTinterwordspacing
W.~Peng, X.~Hong, H.~Chen, and G.~Zhao, ``Learning graph convolutional network
  for skeleton-based human action recognition by neural searching,'' in
  \emph{The Thirty-Fourth {AAAI} Conference on Artificial Intelligence, {AAAI}
  2020, The Thirty-Second Innovative Applications of Artificial Intelligence
  Conference, {IAAI} 2020, The Tenth {AAAI} Symposium on Educational Advances
  in Artificial Intelligence, {EAAI} 2020, New York, NY, USA, February 7-12,
  2020}.\hskip 1em plus 0.5em minus 0.4em\relax {AAAI} Press, 2020, pp.
  2669--2676. [Online]. Available:
  \url{https://aaai.org/ojs/index.php/AAAI/article/view/5652}
\BIBentrySTDinterwordspacing

\bibitem{DBLP:conf/cvpr/Cheng0HC0L20}
\BIBentryALTinterwordspacing
K.~Cheng, Y.~Zhang, X.~He, W.~Chen, J.~Cheng, and H.~Lu, ``Skeleton-based
  action recognition with shift graph convolutional network,'' in \emph{2020
  {IEEE/CVF} Conference on Computer Vision and Pattern Recognition, {CVPR}
  2020, Seattle, WA, USA, June 13-19, 2020}.\hskip 1em plus 0.5em minus
  0.4em\relax Computer Vision Foundation / {IEEE}, 2020, pp. 180--189.
  [Online]. Available:
  \url{https://openaccess.thecvf.com/content\_CVPR\_2020/html/Cheng\_Skeleton-Based\_Action\_Recognition\_With\_Shift\_Graph\_Convolutional\_Network\_CVPR\_2020\_paper.html}
\BIBentrySTDinterwordspacing

\bibitem{DBLP:conf/cvpr/0005LZXXZ20}
\BIBentryALTinterwordspacing
P.~Zhang, C.~Lan, W.~Zeng, J.~Xing, J.~Xue, and N.~Zheng, ``Semantics-guided
  neural networks for efficient skeleton-based human action recognition,'' in
  \emph{2020 {IEEE/CVF} Conference on Computer Vision and Pattern Recognition,
  {CVPR} 2020, Seattle, WA, USA, June 13-19, 2020}.\hskip 1em plus 0.5em minus
  0.4em\relax Computer Vision Foundation / {IEEE}, 2020, pp. 1109--1118.
  [Online]. Available:
  \url{https://openaccess.thecvf.com/content\_CVPR\_2020/html/Zhang\_Semantics-Guided\_Neural\_Networks\_for\_Efficient\_Skeleton-Based\_Human\_Action\_Recognition\_CVPR\_2020\_paper.html}
\BIBentrySTDinterwordspacing

\bibitem{DBLP:conf/mm/Song0SW20}
\BIBentryALTinterwordspacing
Y.~Song, Z.~Zhang, C.~Shan, and L.~Wang, ``Stronger, faster and more
  explainable: {A} graph convolutional baseline for skeleton-based action
  recognition,'' in \emph{{MM} '20: The 28th {ACM} International Conference on
  Multimedia, Virtual Event / Seattle, WA, USA, October 12-16, 2020}, C.~W.
  Chen, R.~Cucchiara, X.~Hua, G.~Qi, E.~Ricci, Z.~Zhang, and R.~Zimmermann,
  Eds.\hskip 1em plus 0.5em minus 0.4em\relax {ACM}, 2020, pp. 1625--1633.
  [Online]. Available: \url{https://doi.org/10.1145/3394171.3413802}
\BIBentrySTDinterwordspacing

\bibitem{DBLP:conf/wacv/ChoM0F20}
\BIBentryALTinterwordspacing
S.~Cho, M.~H. Maqbool, F.~Liu, and H.~Foroosh, ``Self-attention network for
  skeleton-based human action recognition,'' in \emph{{IEEE} Winter Conference
  on Applications of Computer Vision, {WACV} 2020, Snowmass Village, CO, USA,
  March 1-5, 2020}.\hskip 1em plus 0.5em minus 0.4em\relax {IEEE}, 2020, pp.
  624--633. [Online]. Available:
  \url{https://doi.org/10.1109/WACV45572.2020.9093639}
\BIBentrySTDinterwordspacing

\bibitem{liu2019ntu}
J.~Liu, A.~Shahroudy, M.~Perez, G.~Wang, L.-Y. Duan, and A.~C. Kot, ``Ntu rgb+
  d 120: A large-scale benchmark for 3d human activity understanding,''
  \emph{IEEE transactions on pattern analysis and machine intelligence},
  vol.~42, no.~10, pp. 2684--2701, 2019.

\bibitem{li2019spatio}
B.~Li, X.~Li, Z.~Zhang, and F.~Wu, ``Spatio-temporal graph routing for
  skeleton-based action recognition,'' in \emph{Proceedings of the AAAI
  Conference on Artificial Intelligence}, vol.~33, no.~01, 2019, pp.
  8561--8568.

\end{thebibliography}
\end{document}